\documentclass[10pt,journal,compsoc]{IEEEtran}
\ifCLASSOPTIONcompsoc
  \usepackage[nocompress]{cite}
\else
  \usepackage{cite}
\fi
\ifCLASSINFOpdf
\else
\fi

\usepackage{times}
\usepackage{epsfig}
\usepackage{graphicx}
\usepackage{amsmath}
\usepackage{amssymb}
\usepackage{url}
\usepackage{algorithm}
\usepackage{algorithmicx}
\usepackage{enumitem}
\usepackage[noend]{algpseudocode}

\usepackage{xcolor}
\newcommand{\beginsupplement}{%
	\setcounter{table}{0}
	\renewcommand{\thetable}{S\arabic{table}}%
	\setcounter{figure}{0}
	\renewcommand{\thefigure}{S\arabic{figure}}%
	\setcounter{section}{0}
	\renewcommand{\thesection}{S\arabic{section}}
}

\usepackage{comment}
\usepackage{amsfonts}

\begin{document}
%
\title{Canonical Saliency Maps: Decoding Deep Face Models}

%
%
%
%

\author{Thrupthi~Ann~John,
        Vineeth~N~Balasubramanian,~\IEEEmembership{Senior Member,~IEEE,}
        and~C~V~Jawahar,~\IEEEmembership{Member,~IEEE}
\IEEEcompsocitemizethanks{\IEEEcompsocthanksitem Thrupthi Ann John  is and C V Jawahar are with the International Institute of Information Technology, Hyderabad \protect\\
E-mail: thrupthi.ann@research.iiit.ac.in 
\IEEEcompsocthanksitem Vineeth N Balasubramanian is with the Indian Institute of Technology, Hyderabad. \protect\\

}
}
%
%

\markboth{IEEE Transactions on Biometrics, Behavior, and Identity Science (T-BIOM)}%
{John \MakeLowercase{\textit{et al.}}: Understanding Deep Face Models through Canonical Saliency Maps}
%



\IEEEtitleabstractindextext{%
\begin{abstract}
As Deep Neural Network models for face processing tasks approach human-like performance, their deployment in critical applications such as law enforcement and access control has seen an upswing, where any failure may have far-reaching consequences. We need methods to build trust in deployed systems by making their working as transparent as possible. Existing visualization algorithms are designed for object recognition and do not give insightful results when applied` to the face domain. In this work, we present `Canonical Saliency Maps', a new method which highlights relevant facial areas by projecting saliency maps onto a canonical face model. We present two kinds of Canonical Saliency Maps: image-level maps and model-level maps. Image-level maps highlight facial features responsible for the decision made by a deep face model on a given image, thus helping to understand how a DNN made a prediction on the image. Model-level maps provide an understanding of what the entire DNN model focuses on in each task, and thus can be used to detect biases in the model. Our qualitative and quantitative results show the usefulness of the proposed canonical saliency maps, which can be used on any deep face model regardless of the architecture.


\end{abstract}

}

\maketitle

\IEEEdisplaynontitleabstractindextext

%
\IEEEpeerreviewmaketitle

\newcommand{\FIGlfwquant}{
\begin{figure*}[!t]
    \centering
    \includegraphics[width=\linewidth]{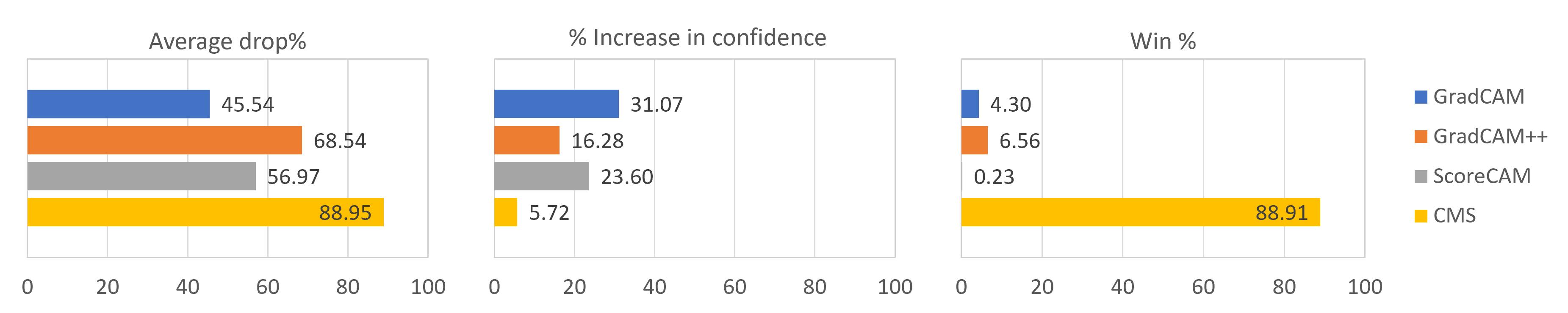}
    \caption{Results for Average Drop \%, \% Increase in Confidence and Win \% of the explanations generated by Grad-CAM, Grad-CAM++, ScoreCAM and CMS on LFW for the VGG-16 model.}
    \label{fig:lfwquant}
\end{figure*}
}

\newcommand{\FIGnetquant}{
\begin{figure}[!t]
    \centering
    \includegraphics[width=\linewidth]{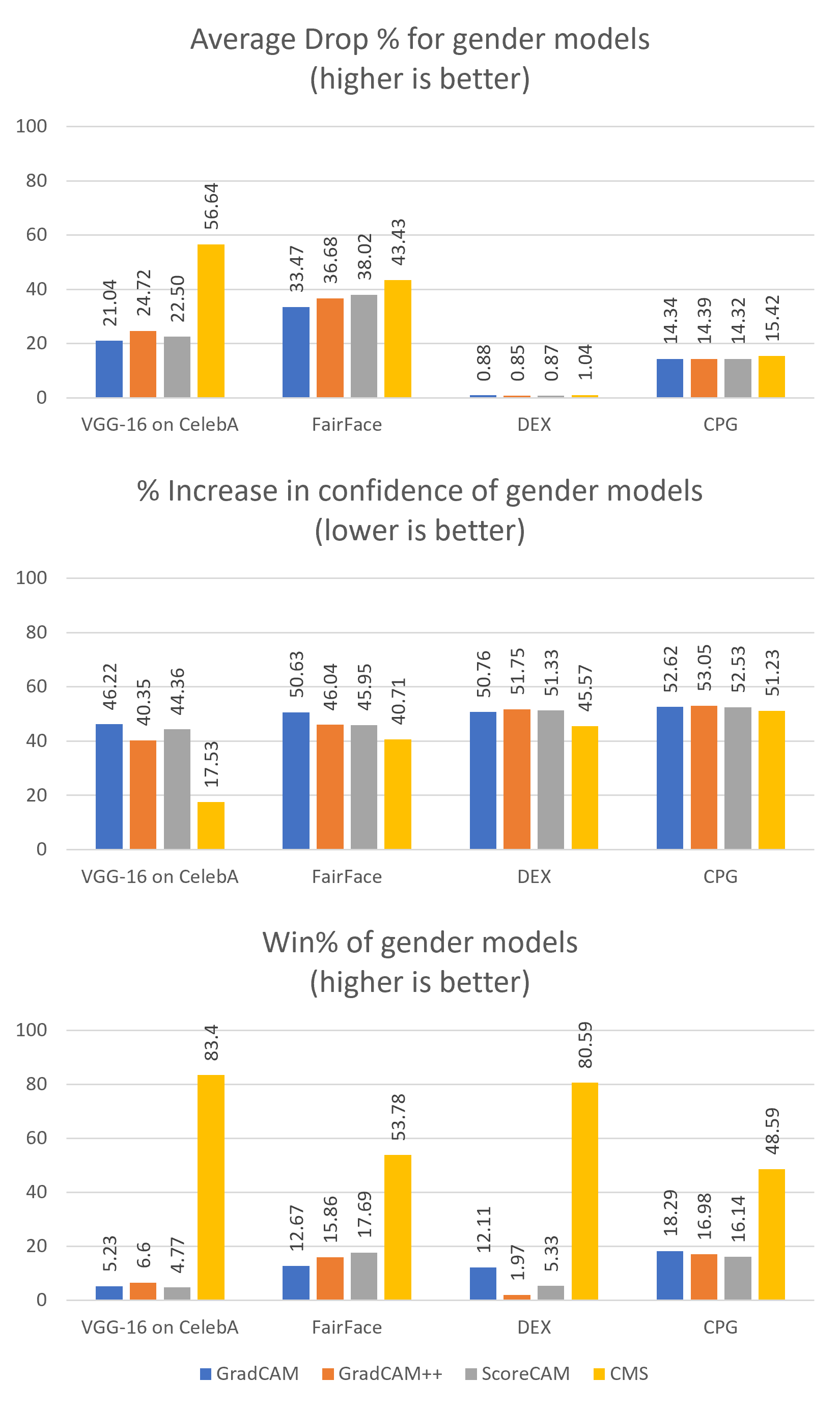}
    \caption{Results for Average Drop \%, \% Increase in Confidence and Win \% of the explanations generated by Grad-CAM, Grad-CAM++, ScoreCAM and CMS on CelebA for various deep face gender models}
    \label{fig:netquant}
\end{figure}
}

\newcommand{\FIGVGGFaceQuant}{
	\begin{figure}[!t]
		\centering
		\includegraphics[width=\linewidth]{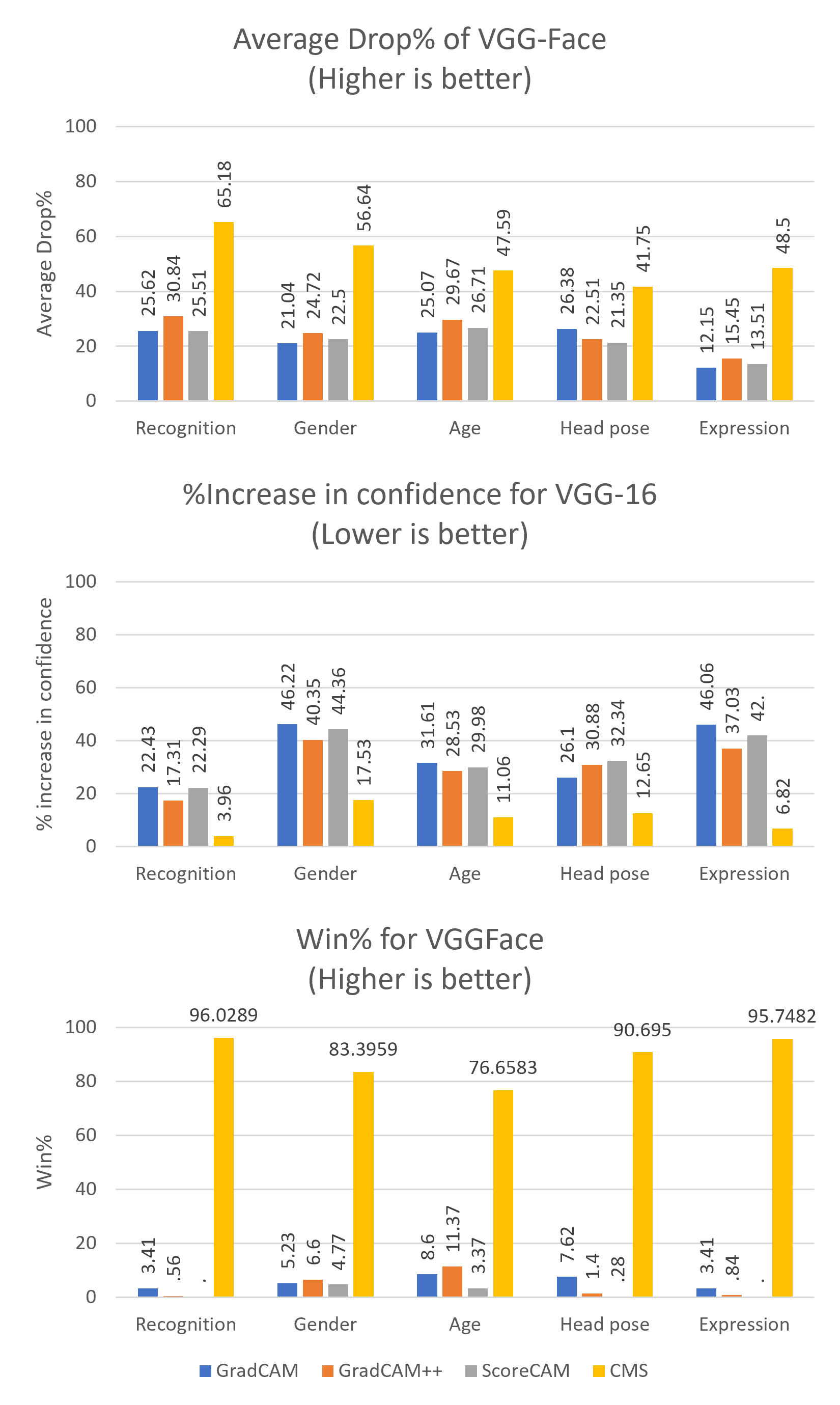}
		\caption{Results for Average Drop \%, \% Increase in Confidence and Win \% of VGG-16 on Celeb-A for the tasks of recognition, gender, age, head pose and expression.}
		\label{fig:vggfacequant}
	\end{figure}
}

\newcommand{\FIGSurvey}{
	\begin{figure}[!t]
		\centering
		\includegraphics[width=0.9\linewidth]{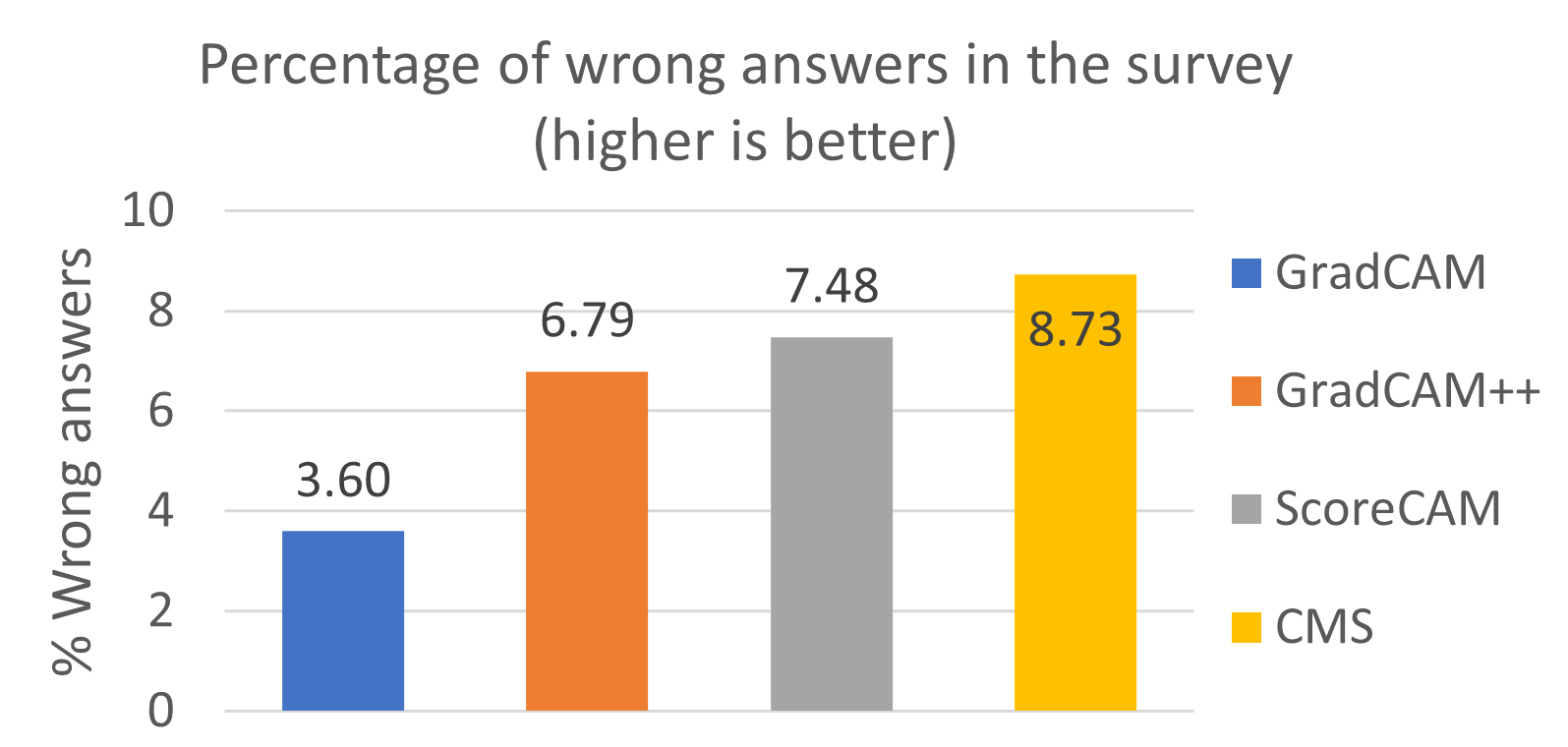}
		\caption{Results for user survey on the perception of gender and emotion on explanation maps. We used 12 base images modified using GradCAM, GradCAM++, ScoreCAM and CMS. The users had to pick binary labels for each image (male-female, happy-sad). Each question was answered by 143 people who were not involved in this project}
		\label{fig:survey}
	\end{figure}
}


\newcommand{\FIGsurveysamples}{
	\begin{figure}[!t]
		\centering
		\includegraphics[width=\linewidth]{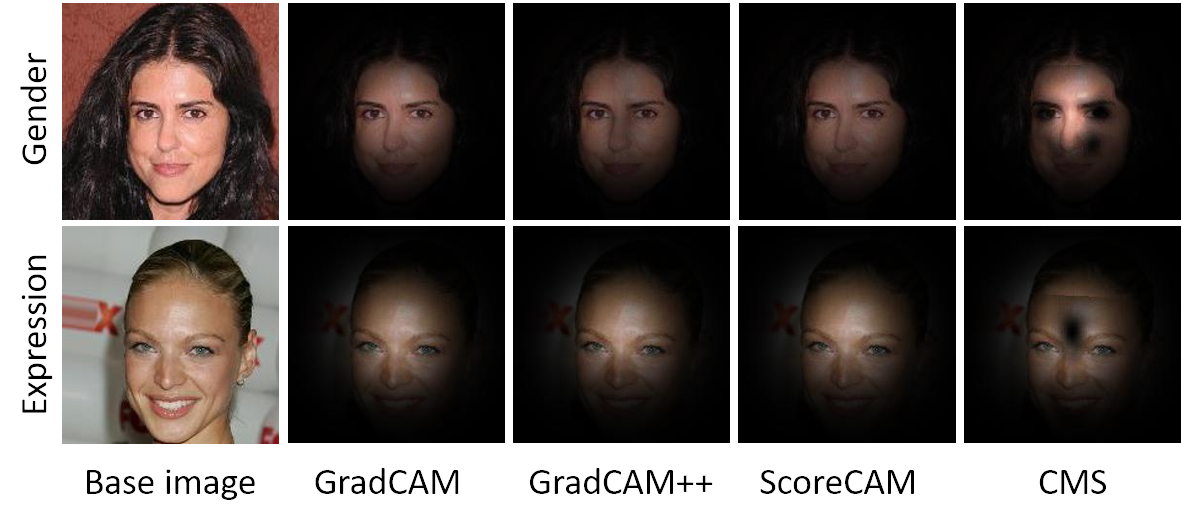}
		\caption{Samples of figures used in our survey (see Section \ref{sec:survey}}
		\label{fig:survey_samples}
	\end{figure}
}

\newcommand{\FIGsurveybase}{
	\begin{figure}[!t]
		\centering
		\includegraphics[width=\linewidth]{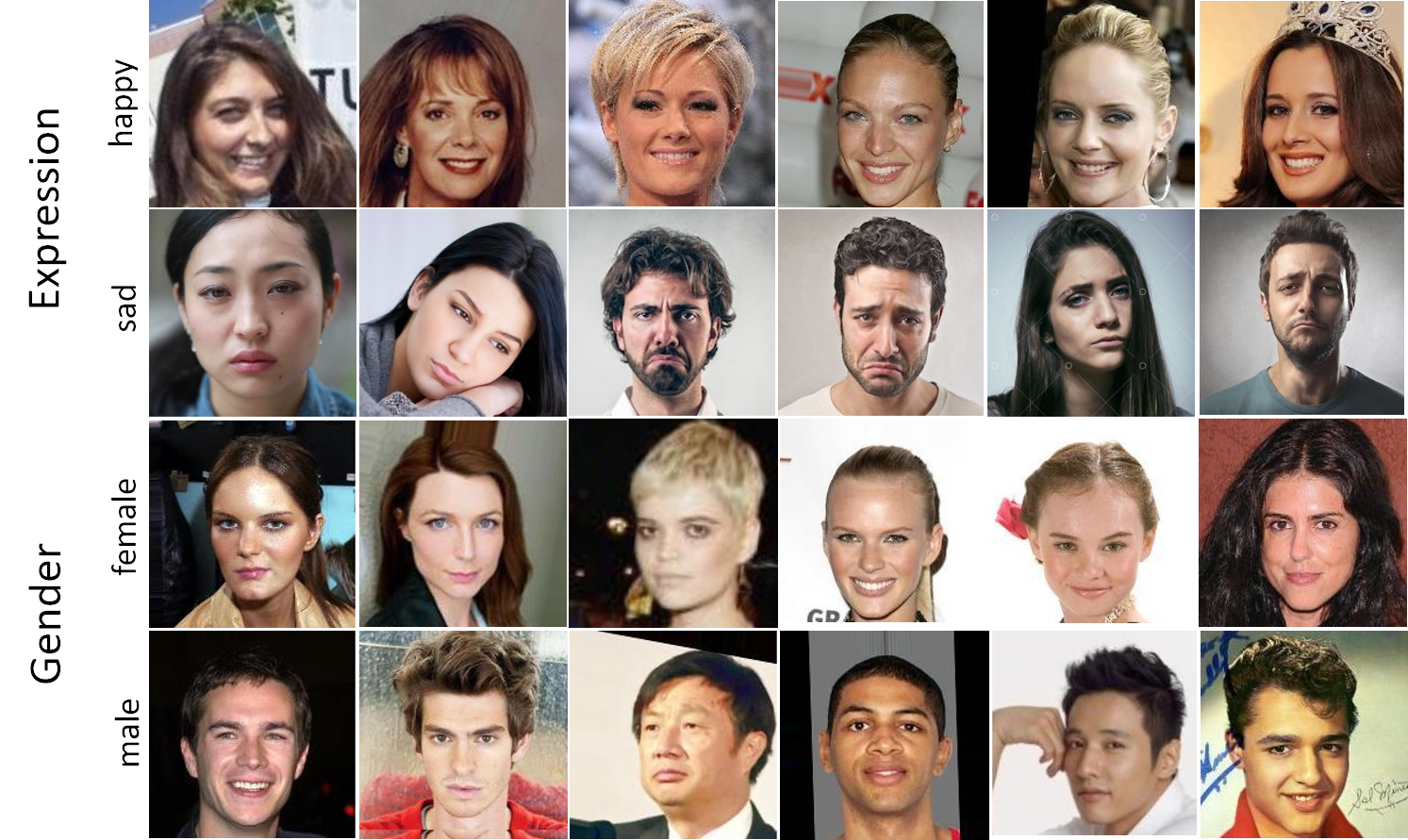}
		\caption{All base images used for our user survey (see Section \ref{sec:survey}}
		\label{fig:survey_base}
	\end{figure}
}

\newcommand{\FIGdensitynorm}{
	\begin{figure}[!t]
		\centering
		\includegraphics[width=0.7\linewidth]{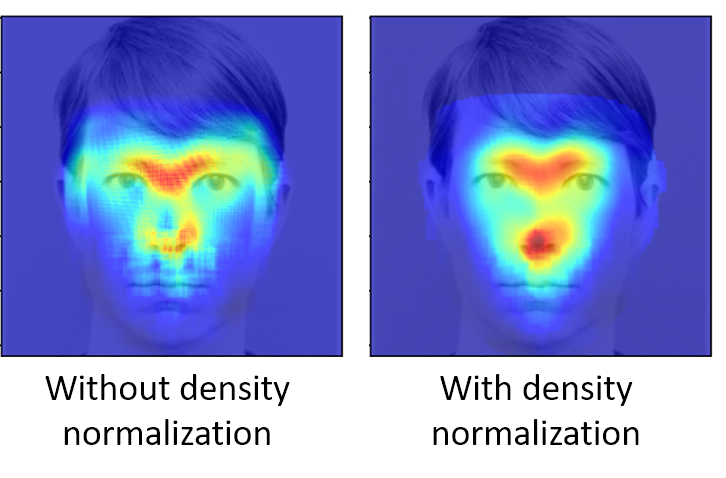}
		\vspace{-12pt}
		\caption{Effect of applying density normalization to the heatmap. Without density normalization, the nose is not highlighted despite it being a discriminative feature, mainly because the density of points on the nose is low}
		\label{fig:densitynorm}
	\end{figure}
}

\newcommand{\FIGalignmentablation}{
\begin{figure*}[!t]
    \centering
    \includegraphics[width=\linewidth]{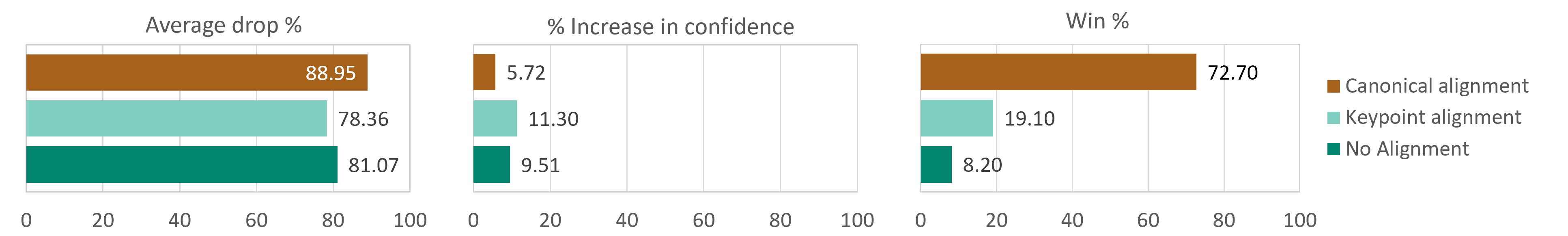}
    \caption{Ablation study on the effect of different types of alignment. Shown are the Average Drop\%, \% Increase in confidence and Win \% for three different types of alignment on the LFW dataset: 1. Canonical face 2. Keypoint-based alignment 3. No alignment.}
    \label{fig:alignment_ablation}
\end{figure*}
}

\newcommand{\FIGalignmentqual}{
\begin{figure}[!t]
    \centering
    \includegraphics[width=\linewidth]{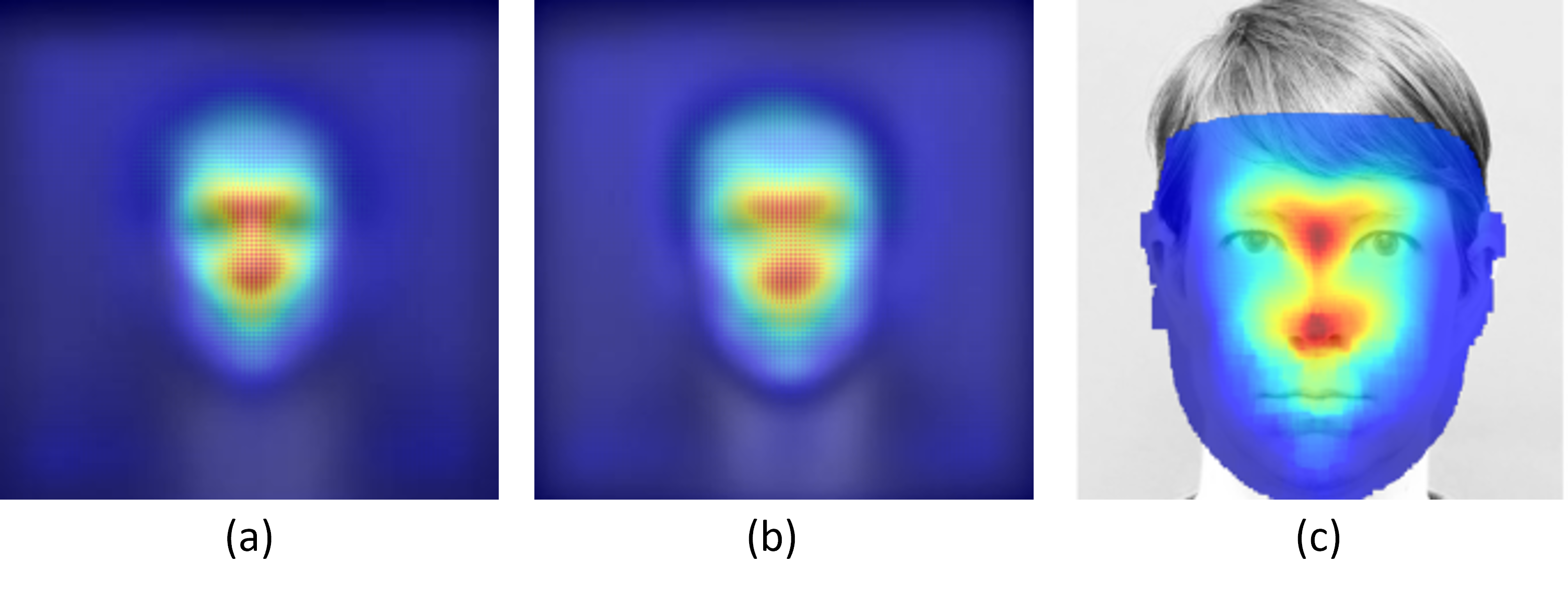}
    \caption{Ablation study on the effect of different types of alignment. Shown are the model saliency maps for three different types of alignment on the LFW dataset: (a) No alignment, superimposed on the average image of LFW; (b) Keypoint-based alignment, superimposed on the average image of LFW-funneled; and (c) CMS superimposed on the canonical face.}
    \label{fig:alignment_qual}
\end{figure}
}

\newcommand{\FIGcmsablation}{
\begin{figure*}[!t]
    \centering
    \includegraphics[width=0.9\linewidth]{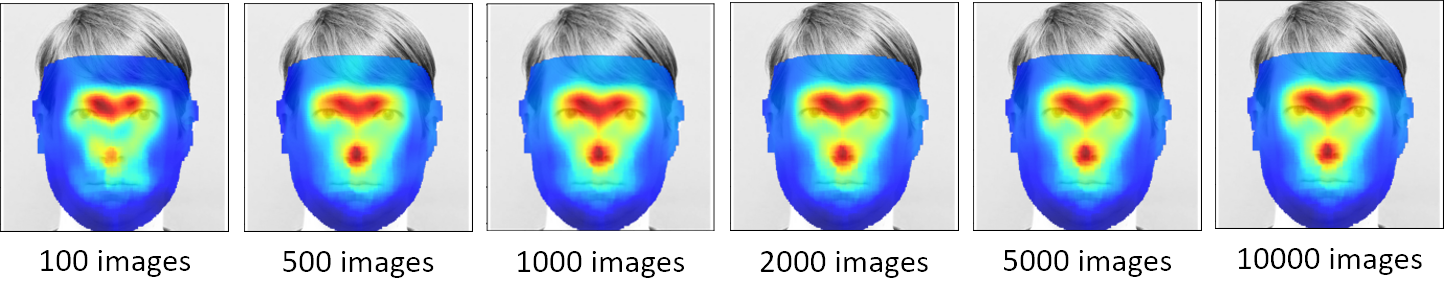}
    \vspace{-10pt}
    \caption{Ablation study to study the effect of the number of images used to create a CMS map. CMS maps for recognition using 100, 500, 1000, 2000, 5000 and 10000 random CIS maps}
    \label{fig:cms_ablation}
\end{figure*}
}

\newcommand{\FIGsizeablation}{
	\begin{figure*}[!ht]
		\centering
		\includegraphics[width=0.9\linewidth]{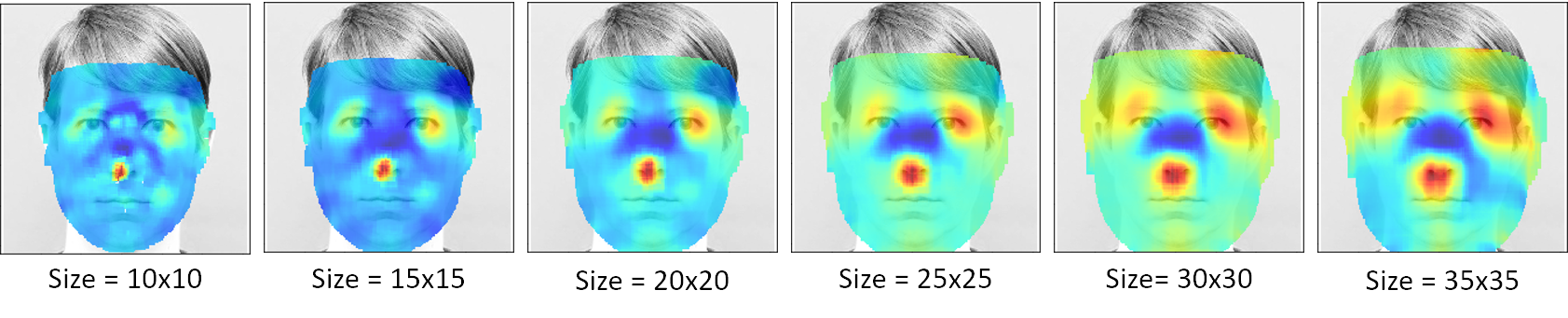}
		\vspace{-12pt}
		\caption{Canonical Image Saliency maps generated when the size of the occluding patch is varied. We used a patch size of $15\times 15$ in all other experiments in this work}
		\label{fig:size_ablation}
	\end{figure*}
}


\newcommand{\FIGmetricjust}{
\begin{figure}[!t]
    \centering
    \includegraphics[width = \linewidth]{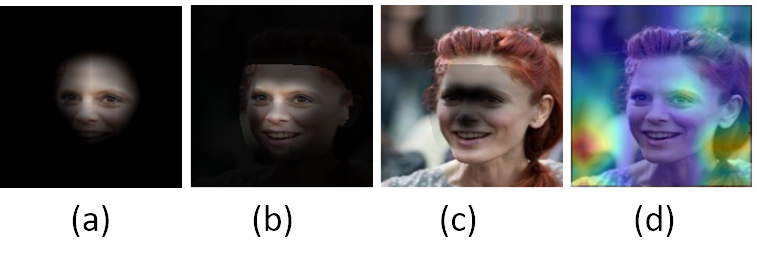}
    \caption{Since face models are trained to look holistically at the face, they have more confidence in figure (a) than in figure (b), even though figure (b) highlights more relevant features. Thus, we use negative saliency maps where darkening relevant features should cause a larger drop in confidence. This also ensures that there is enough context for the model to interpret the face holistically. Another reason for using negative saliency maps is to take care of cases where a visualization method does not interpret the face correctly, as in figure (d). Here, the heatmap completely misses the face and is focused on disparate parts of the image. Using normal explanation maps will result in almost the original image, which will give a high score in the metrics used. This is avoided by using negative explanation maps and normalizing the sum of pixels}
    \label{fig:metric}
\end{figure}
}

\newcommand{\FIGposenose}{
	
	\begin{figure}[!t]
		\centering
		\includegraphics[width=0.9\linewidth]{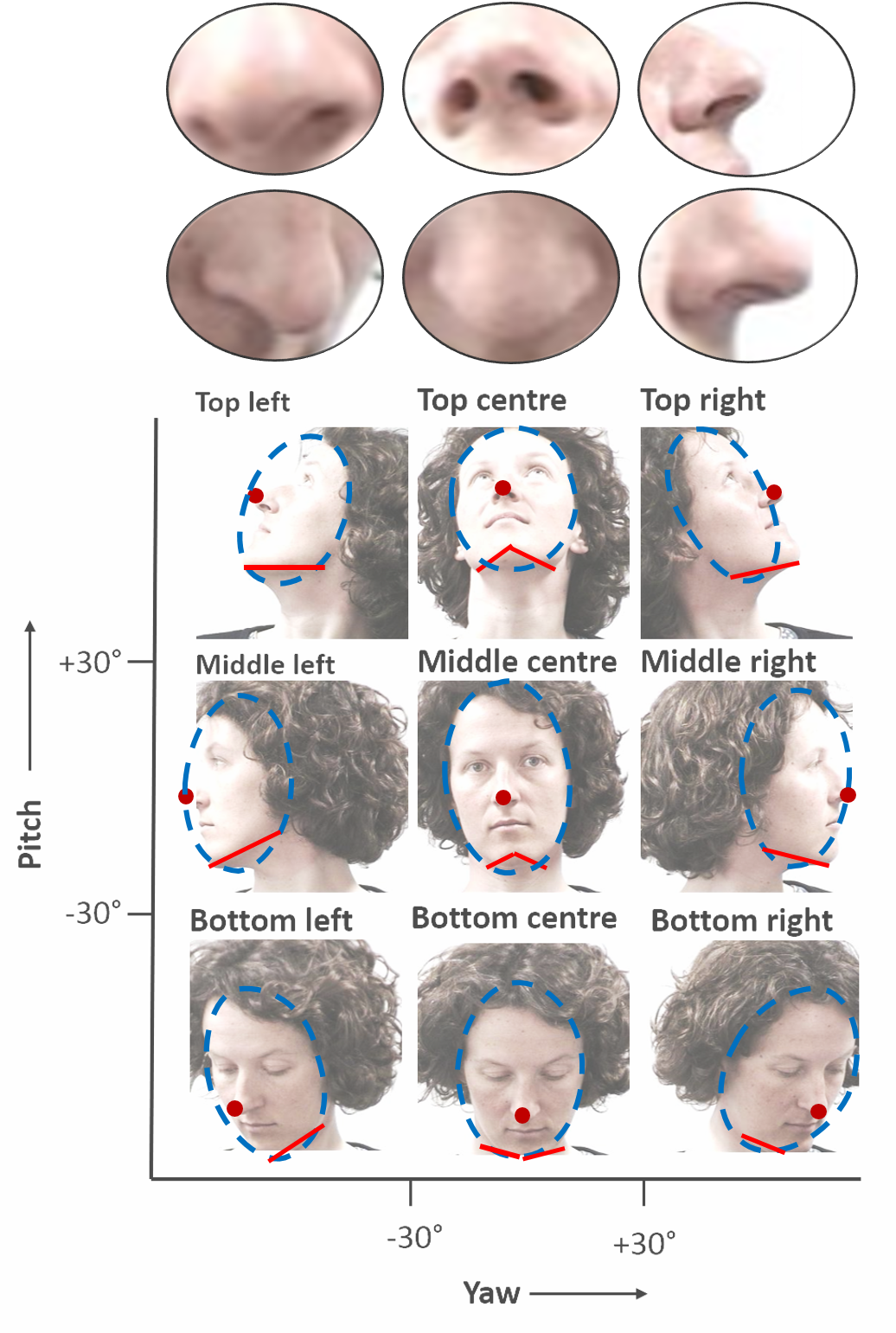}
		\vspace{-15pt}
		\caption{Look at the close-ups of the nose tip in this figure. Can you tell the 3D orientation of the face with this information? The nose, along with jawline, provide a good cue for the face pose. We also observe that the quadrant of the face area in which the nose tip is found is consistent for the same 3D orientation}
		\label{fig:pose_nose}
	\end{figure}
	
}

\newcommand{\FIGvaluealignment}{
	\begin{figure}[!t]
		\centering
		\includegraphics[width=\linewidth]{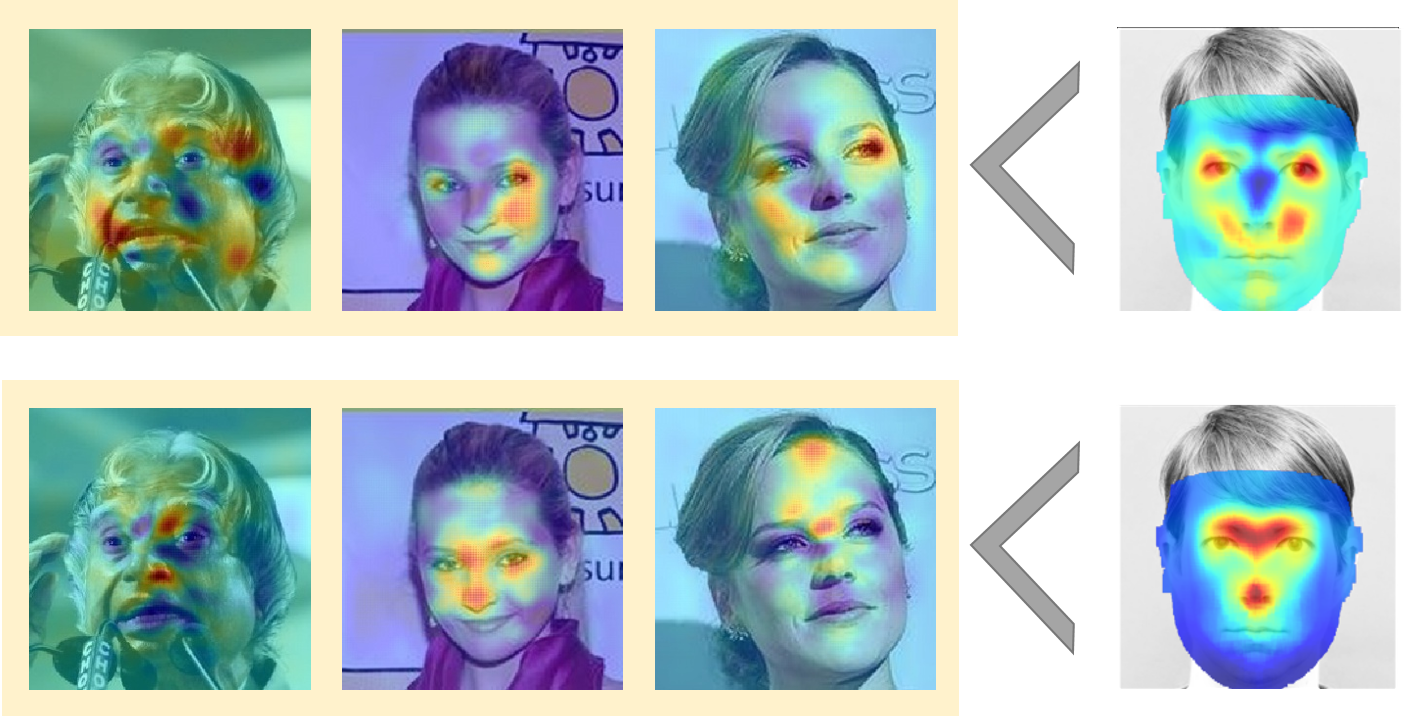}
		\caption{In this figure, we compare individual occlusion maps of gender (first row) and recognition (second row) to the respective cumulative model saliency maps on the right. Individual occlusion maps vary widely and may have slightly different areas highlighted due to differences in pose, occlusion and lighting. Thus, it is hard to compare these images and get the big picture from them. Aggregating heatmaps gets rid of tiny differences caused due to the conditions in which the photo is taken, allowing us to gain valuable insights.}
		\label{fig:value_alignment}
	\end{figure}
}


\newcommand{\FIGqualitative}{
	\begin{figure}[!t]
		\centering
		\includegraphics[width=\linewidth]{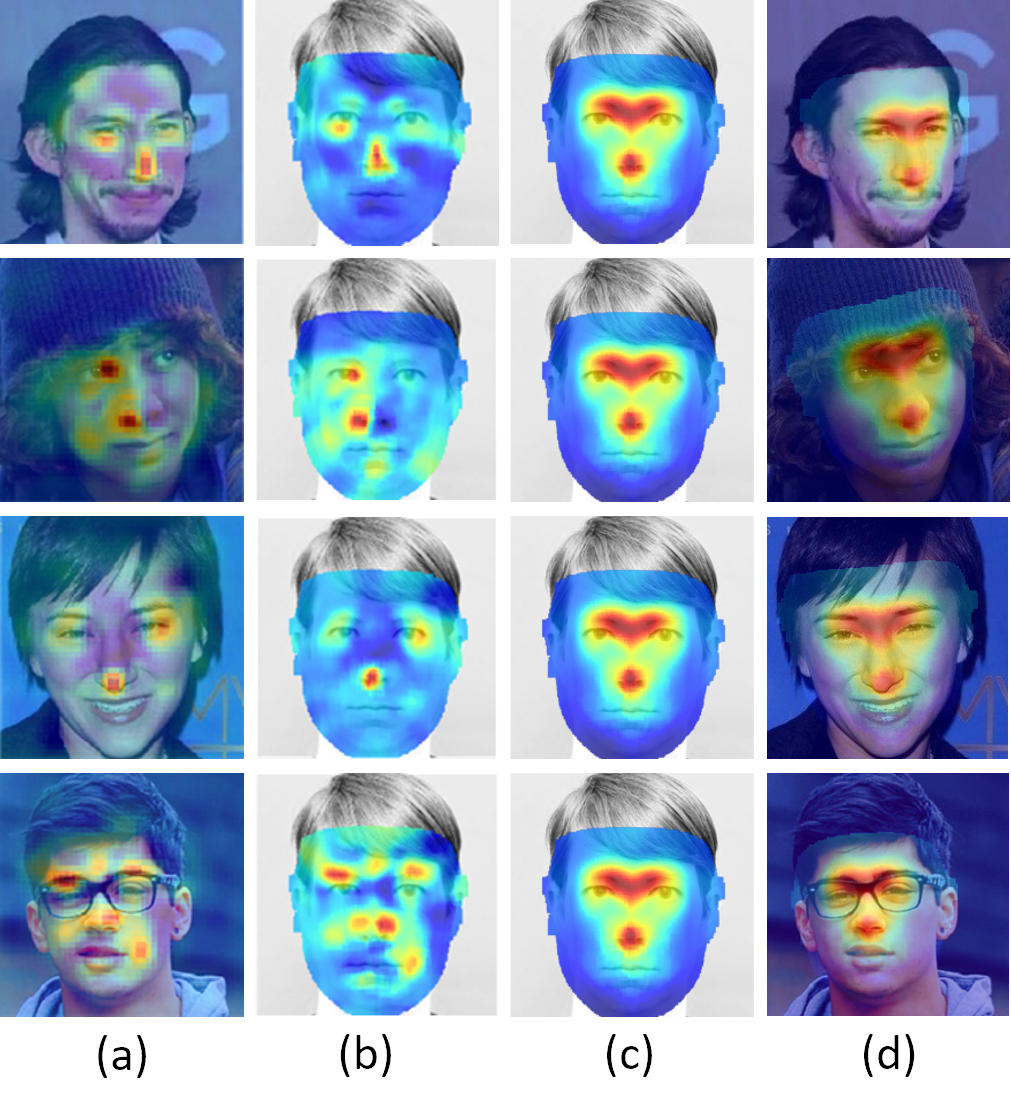}
		\vspace{-15pt}
		\caption{Column (a) shows Occlusion Maps used for saliency visualization (see Section \ref{sec:related}); Column (b) shows Canonical Image Saliency (CIS) maps. CIS maps are a projection of occlusion maps onto a canonical frontal face; Column (c) shows Canonical Model Saliency (CMS) maps. These maps are generated for a model as a whole and hence do not vary with input; Column (d) shows the CMS maps reprojected back onto the input face.}
		\label{fig:qualitative}
	\end{figure}

}

\newcommand{\FIGcms}{
	\begin{figure*}[!t]
		\centering
		\includegraphics[width=0.9\linewidth]{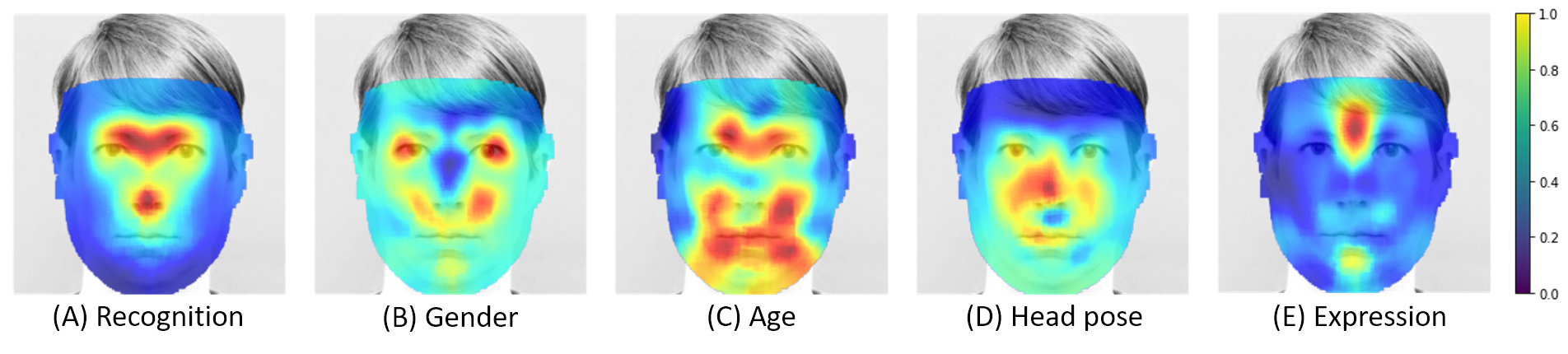}
		\vspace{-6pt}
		\caption{Are all parts of the face of equal importance for different face classification tasks? In this work, we show that deep models do not give equal importance to the entire face. Canonical Model Saliency (CMS) maps show parts of the face that play a significant role in decisions made by the deep model. CMS maps reveal how deep face models work and allow us to detect and diagnose problems inherent to the models, such as biases. For heatmaps, red indicates a high value while blue indicates a low value \textit{(Best viewed in color)}.
		}
		\label{fig:cms}
	\end{figure*}
}

\newcommand{\FIGnetcms}{
	\begin{figure}[!t]
		\centering
		\includegraphics[width=\linewidth]{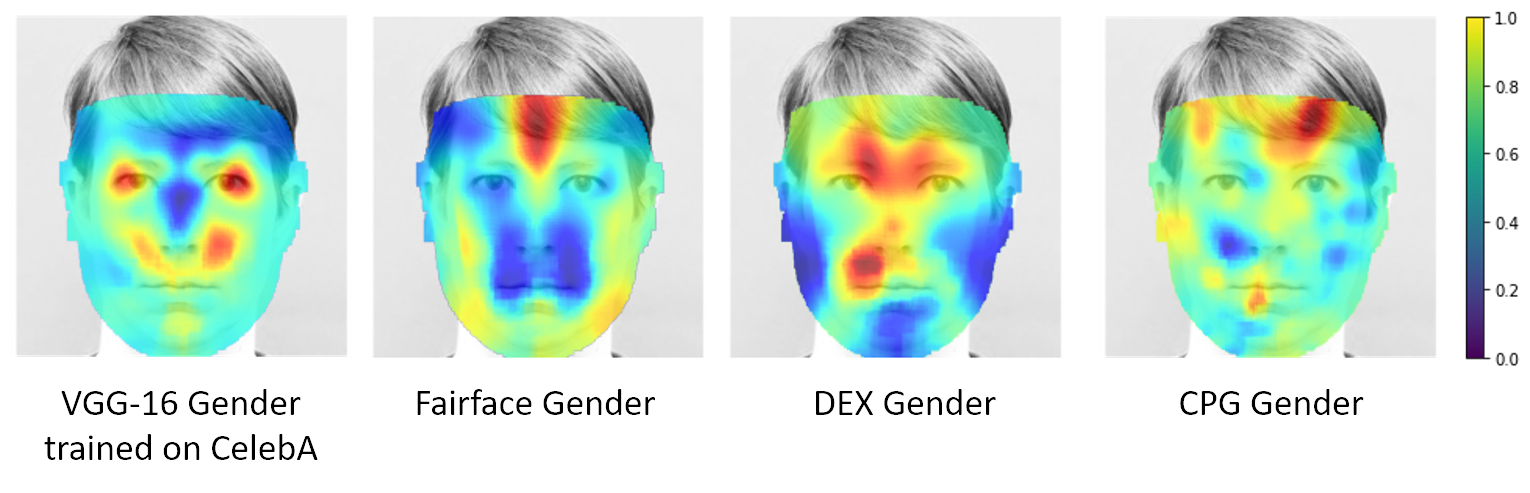}

		\caption{We compare CMS maps obtained from various off-the-shelf deep gender models}
		\label{fig:netcms}
	\end{figure}
}

\newcommand{\FIGnoclassification}{
\begin{figure}
    \centering
    \includegraphics[width=0.6\linewidth]{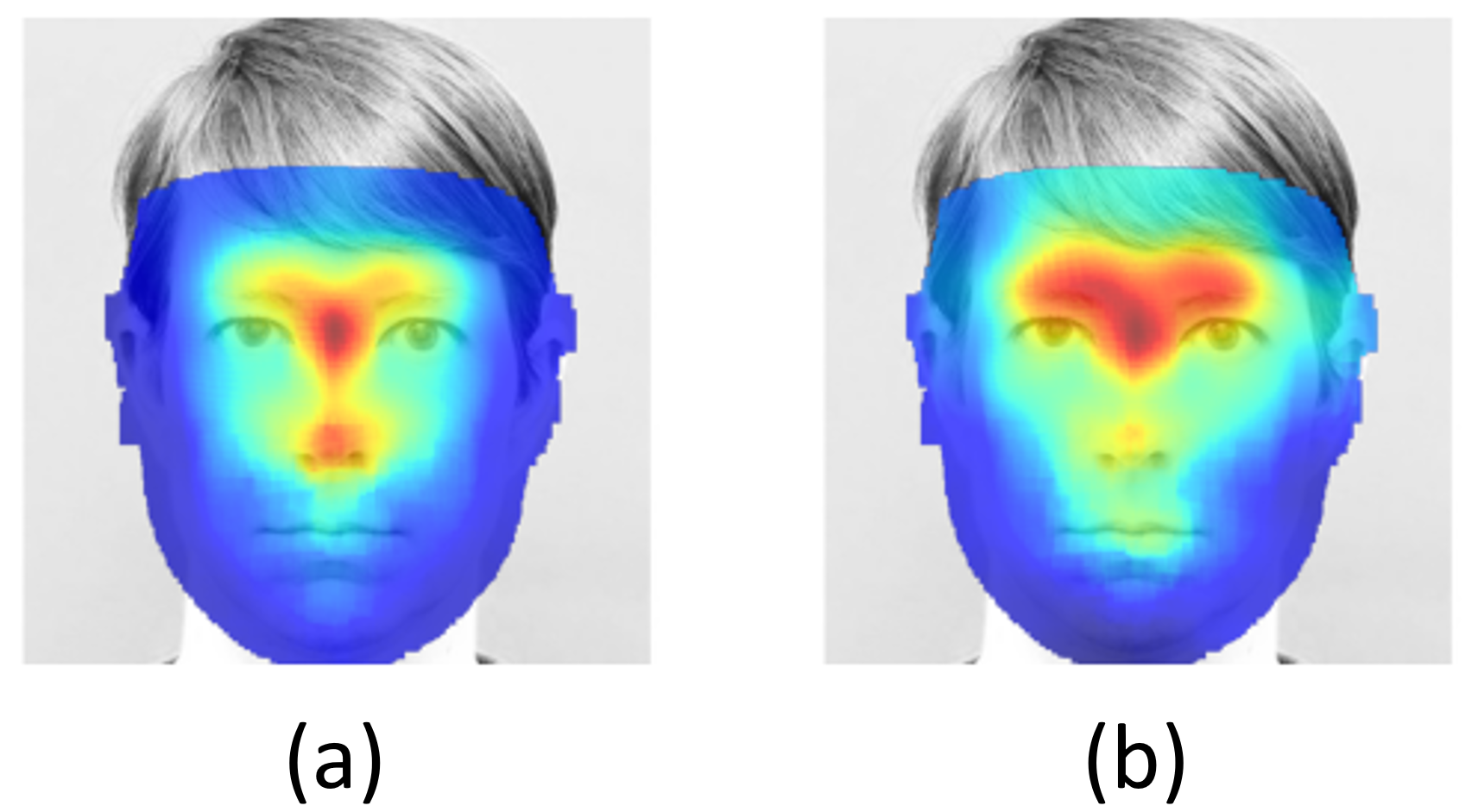}
    \caption{Calculating CMS maps for non-classification tasks on the LFW dataset: (a) CMS map for zero-shot learning of identity using nearest neighbour; (b) CMS map for face verification}
    \label{fig:noclassification}
\end{figure}
}

\newcommand{\FIGsanitycheck}{
	\begin{figure}[!t]
		\centering
		\includegraphics[width=\linewidth]{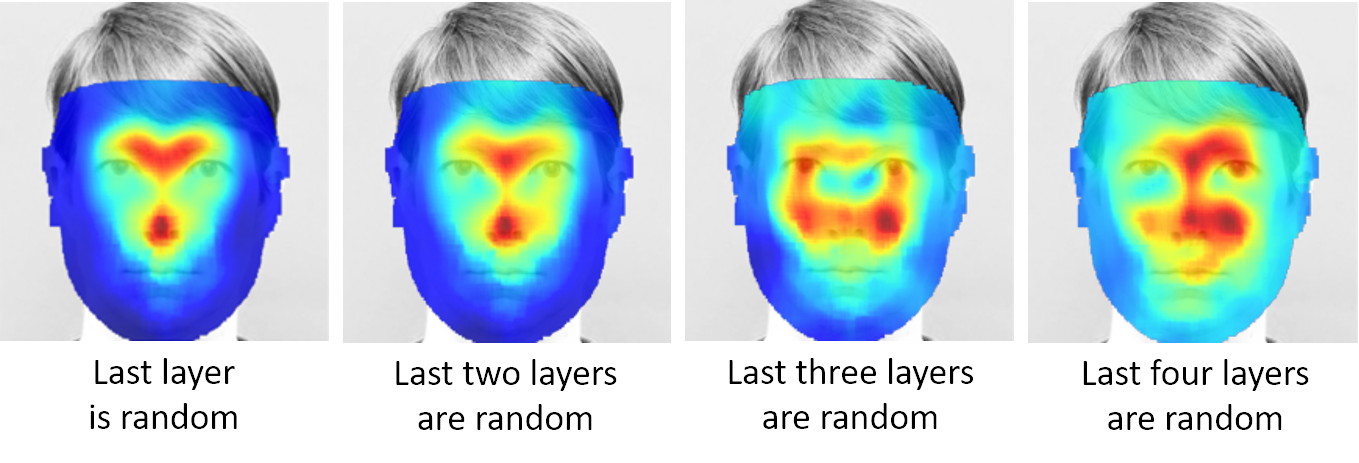}
		\caption{Sanity check on our visualization method. We progressively randomized the layers of the VGG-16 face model starting with the output layer as described in \cite{sanitycheck}. We observe that the CMS map gets progressively randomized; our method passes the sanity check. (a) Last layer randomized; (b) Last two layers randomized; (c) Last three layers randomized; (d) Last four layers randomized}
		\label{fig:sanity}
	\end{figure}
}

\newcommand{\FIGsaliencyrelated}{
	\begin{figure*}[!t]
		\centering
		\includegraphics[width=0.95\linewidth]{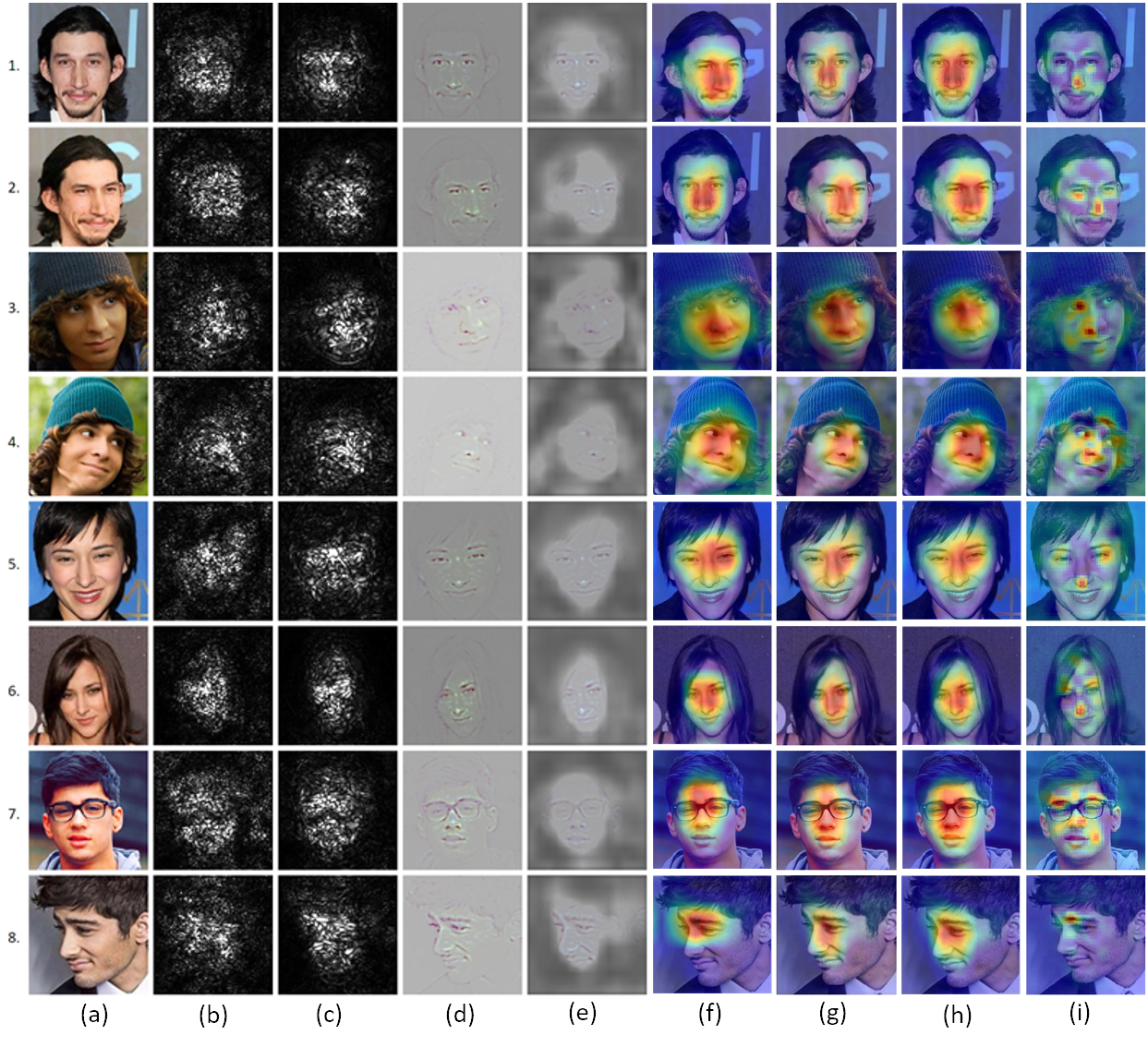}
		\vspace{-12pt}
		\caption{A comparison of various saliency visualization methods on the VGG-Face model \cite{VGGFace} for the task of face recognition. For each image, the target class of the visualization is the ground truth class. (a) Original image; (b) Vanilla gradients\cite{vanillagradient}; (c) Smooth-grad \cite{smoothgrad}; (d) Guided Backpropagation \cite{guidedbackprop}; (e) Guided GradCAM++ \cite{gradcam++}; (f) GradCAM \cite{gradcam}; (g) GradCAM++ \cite{gradcam++}; (h) ScoreCAM \cite{scorecam}; (i) Occlusion map \cite{deconvnet}. Images are taken from the VGG-Face dataset \cite{VGGFace}. Rows (1, 2), (3, 4), (5, 6) and (7, 8) have the same identity. (Best viewed in color)}
		\vspace{-8pt}
		\label{fig:saliency_related}
	\end{figure*}
}


\newcommand{\FIGcisprocedure}{
	\begin{figure}[!t]
		\centering
		\includegraphics[width=\linewidth]{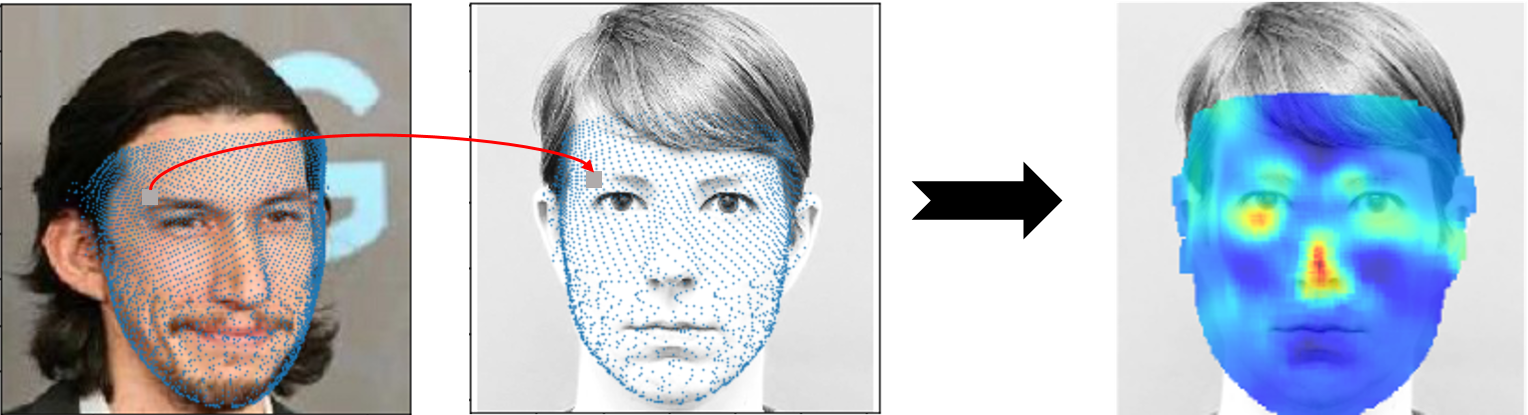}
		\caption{Procedure of computing Canonical Image Saliency (CIS) map. First, the input face is densely aligned. Each part of the input face is occluded with a small patch and the classification confidence is obtained. The drop in confidence is plotted on the same face location on a neutral face image to obtain the Canonical Image Saliency map}
		\label{fig:cis_procedure}
	\end{figure}
}

\newcommand{\ALGcis}{
	\begin{algorithm}
		\DontPrintSemicolon
		\KwIn{
			\begin{itemize}
				\item input image $I$ of size $u\times u$ \\
				\item $input\_mesh$ of size $n\times 2$ (n 2-dimensional points)\\
				\item $frontal\_image$ of size $v\times v\times 3$ \\
				\item $frontal\_mesh$ of size $n\times 2$ \\
				\item $model$: deep model to find saliency where $model(I)$ gives a vector of class probabilities \\
				\item $target\_class$ of the input image I \\
				\item $sz$: size of occlusion square
			\end{itemize}
		}
		\KwOut{heatmap}
		$heatmap$ $\gets$ zero matrix of size $u\times u$ \\
		$count$ $\gets$ zero matrix  of size $u\times u$ \\
		$fsz \gets fsz\times \frac{v}{u}$ \tcp{normalize square size} 
		$original$ $\gets$ $model(input\_image)[target\_class]$ \tcp{Get the classifier confidence}
		\For{$i \gets 0$ \textbf{to} $n$ } {
			
			$x_{I} , y_{I} \gets input\_mesh[i][0], input\_mesh[i][1] $ \\
			
			$I[x_{I}-\frac{sz}{2}:x_{I}+\frac{sz}{2}][y_{I}-\frac{sz}{2}:y_{I}+\frac{sz}{2}] \gets 0 $\\
			
			$difference \gets original - model(I)[target\_class] $\\
			
			$x_{frontal}, y_{frontal} \gets frontal\_mesh[i][0], frontal\_mesh[i][1]$\\
			
			$heatmap[x_{frontal}-\frac{fsz}{2}:x_{frontal}+\frac{fsz}{2}][y_{frontal}-\frac{fsz}{2}:y_{frontal}+\frac{fsz}{2}] += difference $\\
			
			$count[x_{frontal}-\frac{fsz}{2}:x_{frontal}+\frac{fsz}{2}][y_{frontal}-\frac{fsz}{2}:y_{frontal}+\frac{fsz}{2}] += 1$ 
		}
		\tcp{Normalize by count, taking care of divide-by-0}
		$heatmap[count=0] \gets 0 $\\
		$count[count=0] \gets 1 $\\
		$heatmap \gets \frac{heatmap}{count}$ \\
		\Return{heatmap}
		\caption{Canonical Image Saliency map}
		\label{alg:cis}
	\end{algorithm}

}

\newcommand{\ALGcisnew}{
	\begin{algorithm}
		\DontPrintSemicolon
		\KwIn{
			\begin{itemize}
				\item input image $I$ of size $W_I\times H_I$ \\
				\item input mesh $M_I$ of size $N\times 2$ (N 2-dimensional points)\\
				\item frontal image $F$ of size $W_F\times H_F\times 3$ \\
				\item frontal mesh $M_F$ of size $N\times 2$ \\
				\item model $\phi$: deep model to find saliency where $\phi(I,c)$ gives the confidence of I for class c \\
				\item target class $C$ of the input image I \\
				\item $sz$: size of occlusion square
			\end{itemize}
		}
		\KwOut{heatmap $P$ of size $v\times v$}
		$P \gets \{0\}^{W_F\times H_F}$\\
		$N \gets \{0\}^{W_F\times H_F}$ \\
		$fsz \gets fsz\times \frac{H_F}{H_I}$ \tcp{adjust square size} 

		\For{$i \gets 0$ \textbf{to} $n$ } {
			$I^* \gets I$ \\
			$I^*[M_I[i,0]-\frac{sz}{2}:M_I[i,0]+\frac{sz}{2}][M_I[i,1]-\frac{sz}{2}: M_I[i,1]+\frac{sz}{2}] \gets 0 $ \tcp{occlude a small square}
			
			$x_F, y_F \gets M_F[i,0], M_F[i,1]$\\
			
			$P[x_F-\frac{fsz}{2}:x_F+\frac{fsz}{2}][y_F-\frac{fsz}{2}:y_F+\frac{fsz}{2}] +=  \phi(I,C) - \phi(I^*,C) $  \tcp{Assign difference of confidence to heatmap}
			
			$N[x_F-\frac{fsz}{2}:x_F+\frac{fsz}{2}][y_F-\frac{fsz}{2}:y_F+\frac{fsz}{2}] += 1$ 
		}
		\tcp{Normalize by count, taking care of divide-by-0}
		$P[N=0] \gets 0 $\\
		$N[N=0] \gets 1 $\\
		$P \gets P \oslash N $\\ 
		\Return{heatmap}
		\caption{Canonical Image Saliency map}
		\label{alg:cisnew}
	\end{algorithm}

}


\newcommand{\FIGgenderablation}{    
	\begin{figure*}[!t]
		\centering
		\includegraphics[width=\linewidth]{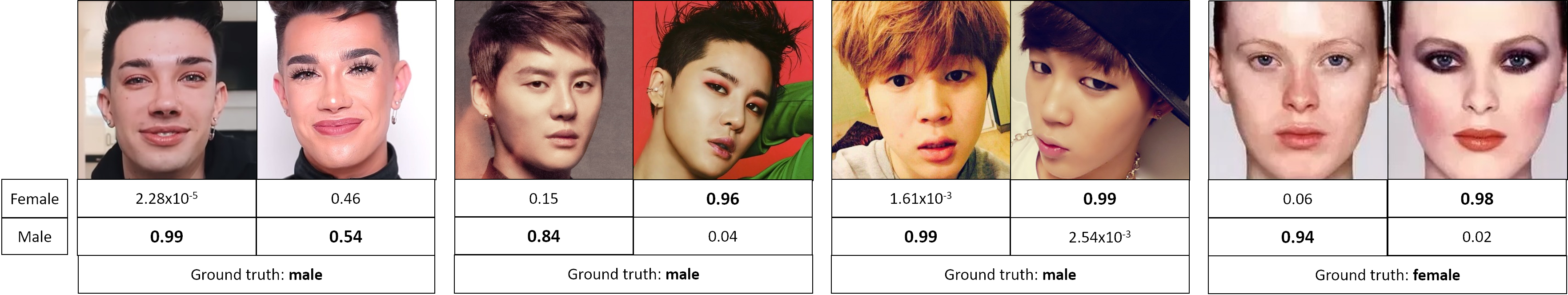}
		\caption{Make-up matters! The figure shows the classification confidence of a gender model on the same person with and without eye make-up. The top row shows the confidence for `female' classification and the bottom row shows the confidence for `male' classification. The ground truth label is given below each pair of images.}
		\label{fig:gender_ablation_big}
	\end{figure*}
}

\newcommand{\FIGgenderabl}{    
	\begin{figure}[!t]
		\centering
		\includegraphics[width=\linewidth]{Images/gender_abl2.png}
		\caption{Make-up matters! Figure shows the classification confidence of a gender model on the same person with and without eye make-up. The top row shows the confidence for `female' classification and the bottom row shows the confidence for `male' classification. The ground truth label is highlighted in green.}
		\label{fig:gender_ablation_small}
	\end{figure}
}

\newcommand{\FIGcmslightcnn}{
	\begin{figure}[!t]
		\centering
		\includegraphics[width=0.5\linewidth]{Images/average_recognition_lightcnn.png}
		\caption{Facial areas that are most important for recognition for a LightCNN \cite{lightcnn} model using its Canonical  Model Saliency map. Compare the output with the Canonical Model Saliency map for the recognition model in Figure \ref{fig:average_heatmaps}A.}
		\label{fig:cms_lightcnn}
	\end{figure}
}

\IEEEraisesectionheading{\section{Introduction}\label{sec:introduction}}
\FIGcms
\IEEEPARstart{D}{eep} learning achieves state-of-the-art performance in most computer vision tasks, surpassing earlier methods by a large margin. The performance of deep neural networks is improving in leaps and bounds for face processing tasks such as face recognition and detection. In 2014, DeepFace \cite{deepface} approached human-like performance for the first time on the LFW benchmark \cite{lfw}, a dataset of face images in unconstrained settings (DeepFace: 97.35\% vs. Human: 97.53\%), using a training dataset of 4 million images. In recent years, the accuracy has increased up to 99.8\% \cite{recognitionsurvey}, thereby surpassing human performance on the benchmark. Deep face models are now deemed to be real-world ready. They are used in many critical areas by government agencies including law enforcement and access control. Currently, models for face tasks are available from major companies like Microsoft, IBM and Amazon who claim that their models are highly accurate. In this scenario, two crucial questions arise: Do pre-trained models perform as well as they claim, and how do we find the weaknesses existing in these models and improve them. 

Failures of face models in critical areas have far-reaching and devastating consequences. Inaccuracies in facial recognition technology can result in an innocent person being misidentified as a criminal and subjected to unwarranted police scrutiny. Big Brother Watch UK released the Face-Off report \cite{BigBrotherUK} highlighting false positive match rates of over 90\% for the facial recognition technology deployed by the Metropolitan police. A recent study \cite{gendershades} demonstrated that although commercial software solutions report high accuracies (Amazon's Rekognition reports an accuracy of 97\%), they demonstrate skin-type and gender biases that go unreported as the benchmarks themselves are skewed. When performance is reported on public or private databases, they are always subject to the biases inherent in these databases. The algorithms may be then used in the real world in conditions that differ wildly from the ones that they are tested in, causing the algorithms to produce erroneous results. How do we catch such issues at an early stage? High reported accuracy is not enough to determine how an algorithm will perform under real-life conditions. We need to be able to peek inside the algorithms and understand how they work. The opaqueness of deep models restricts its usefulness in highly regulated environments (e.g. healthcare, autonomous driving), which may require the reasoning of the decisions taken by the deep models to be provided. To build trust in deployed intelligent systems, they need to be transparent i.e. they should be able to explain why they predict what they predict \cite{gradcam}. Interpretable algorithms allow us to responsibly deploy deep face models in the real world, as they help end users be aware of these models' characteristics and shortcomings. 

Several visualization methods have been proposed to increase the interpretability and transparency of deep neural networks. So far, most neural network visualization methods have been created with the task of object recognition in mind. There have been very few works that applied these algorithms exclusively to the face domain \cite{fgvisualization, zhong2018deep}. The saliency methods of object recognition do not readily translate to the face domain, as the images used for face tasks have different properties from those used for generic object recognition. Face images are highly structured forms of input. The intra-class difference is very small and face tasks are a form of fine-grained classification. Input images  to face classification models are usually pre-processed so that they are centered around the face of interest and there is only one face per image. Examples of current saliency methods applied to faces are given in Figure \ref{fig:saliency_related}. We observe that most methods highlight a large area in the center of the face. This type of heatmap may be useful for object recognition when there are multiple objects in a single image, but shows only trivial information for face images.
Since faces are centered in the input image, the question `where in the image' is not as relevant as `where on the face'. In this work, we introduce a simple yet effective `standardization' process for visualization of deep learning models for face processing, that converts image coordinates to face coordinates and thus  makes the resultant saliency maps more effective in practice. We utilize the structure of faces and project the saliency maps onto a standard frontal face to obtain \textit{`Canonical Saliency Maps'} that are independent of image coordinates. These canonical saliency maps can be further processed to compare images or observe trends.

To this end, we propose two types of canonical maps: Canonical Image Saliency (CIS) maps and Canonical Model Saliency (CMS) maps. CIS maps are detailed attention maps of input faces projected onto a standard frontal face. CMS maps, on the other hand, globally visualize the characteristic heatmap of an entire face network, as opposed to an input image. This shows the general trend of facial features a network fixates on while making decisions. Such a model-level saliency map can only be generated using a canonical approach, and not by currently available saliency maps. CMS maps highlight areas that are of most significance for a given face task across a dataset. Since we need only the confidence of the classifier for this purpose, these can be generated for any available model or architecture, even if the implementation details are not available. Thus, this approach may even be used for analyzing commercial models that may not reveal their architecture designs. 
 
In order to validate our contributions comprehensively, we study our canonical maps on five different face processing tasks: face recognition, gender recognition, age recognition, head pose estimation and facial expression recognition (Figure \ref{fig:cms}). We use well-known architectures in our studies and also compare the fixation patterns of the models for human recognition of faces. 
We also show that our visualization method helps discovers a bias in gender recognition models which rely on eye make-up to make decisions. 

Our key contributions can be summarized as follows:
\begin{itemize}
    \item We present a method to standardize face saliency images and project them from image coordinates to face coordinates. This `standardization' produces canonical heat-maps that show the relevance of different facial parts to a deep face task. The new maps are more insightful than the saliency maps produced by current methods and can be used for comparison and observation of trends.
    \item We introduce two types of canonical heatmaps: (i) Canonical Image Saliency maps which highlight the significant facial areas of a specific input image pertinent to a prediction; and (ii) Canonical Model Saliency maps, which capture global characteristics of an entire deep face model while making predictions across data points, which allows us to understand the network and potentially diagnose problems. 
    \item Our algorithms can be performed on any face model even if the implementation is not available. We demonstrate the superior performance of our method using extensive suite of experiments. 
    \item We explore the working of deep face models trained for various face tasks having different architectures. We illustrate how to interpret the canonical maps and demonstrate their diagnostic utility  by detecting a bias that arises from using a celebrity face dataset to train a deep network to classify gender. 
\end{itemize}

\section{Related Work}
\label{sec:related}
\FIGsaliencyrelated
There has been extensive research dedicated to saliency visualization methods in recent years. One of the first efforts to obtain image saliency was by Simonyan et al \cite{vanillagradient} which used the magnitude of the gradients to obtain a noisy and scattered saliency map. Zeiler and Fergus \cite{deconvnet} and Springernberg et al. \cite{guidedbackprop} subsequently introduced methods to highlight the important details of the image. These visualizations were not class-sensitive. Zeiler and Fergus \cite{deconvnet} also proposed a method to obtain coarse class-specific saliency maps by occluding parts of the input image and monitoring the output of the classifier. Recent works such as CAM \cite{cam}, GradCAM \cite{gradcam}, GradCAM++ \cite{gradcam++} and ScoreCAM \cite{scorecam} proposed gradient-based methods to produce coarse, class-sensitive saliency maps that highlights areas of the input image that were influential in the classifier output. Smilkov et al. proposed a technique called `SmoothGrad' \cite{smoothgrad} which produced a smooth version of such maps by averaging gradient maps after perturbing the input image with noise.  

Although there have been many methods introduced for saliency visualization for general image classification settings, such methods do not explicitly address non-trivial fine-grained details when used on face images, as shown in Figure \ref{fig:saliency_related}. Columns (b) and (c) in the figure show results of methods that use the magnitude of gradients to produce a heatmap. These heatmaps are scattered and it is difficult to see the details and interpret classification results using them. Guided backpropagation, shown in column (d), shows the finer details of the face, but is not class-sensitive, thus reducing their utility for interpretation. Columns (f), (g) and (h), corresponding to GradCAM \cite{gradcam}, GradCAM++ \cite{gradcam++} and ScoreCAM \cite{scorecam}, are class-specific, but most commonly highlight the central area of a face making them uninformative across different face processing tasks. Column (e) represents the results of Guided GradCAM++, obtained by multiplying the output of guided backpropagation with the GradCAM++ heatmap, shows fine details while highlighting the class-discriminative area of the face. Occlusion maps in column (i) of Figure \ref{fig:saliency_related} seem to give the most informative results for our use case. This method maps the impact that each region of the image has on the classification, in effect mapping out how representative of the class each region is. It produces a more non-trivial heatmap showing finer details than the other heatmaps. The heatmap resolution can also be adjusted by changing the size of the occlusion and the stride, and the method can be used with any architecture and loss function. Our visualization method is hence built on occlusion maps given this inference from our studies on face images. 
The closest method to ours is \cite{zhong2018deep}, which uses occlusion maps generated between pairs of similar-looking face images to assist humans in telling them apart. They  do this by aligning two faces using keypoints and systematically occluding patches of both faces  and  recording the  change  in cosine  similarity  between  the faces  on  a heatmap. The resulting heatmaps reflect the degree of difference between the face pairs. Unlike this work, our method works on multi-class classification tasks and introduces the face canonicalization procedure. 


Our proposed Canonical Model Saliency Maps visualize saliency of face networks w.r.t. different regions of the face for different face processing tasks. These maps allow us to conduct useful analysis by comparing the facial areas important to the network to the areas that are expected to be important to classify the task. However, the challenge herein is - how do we obtain the `correct' expectations to compare the network's saliency map to? One may look at human cognition as a benchmark for what a deep network should see. 
Extensive research exists on how humans recognize faces; important results have been presented recently in \cite{psychology19}. For e.g., humans are known to be good at recognizing low-resolution and degraded faces, when compared to machines. There is a marked difference in the recognition rate of humans when seeing familiar faces when compared to unknown faces. The face's top part, especially the eyebrows, is known to be an important cue for human face recognition \cite{psychology19}. Comparing our face saliency maps with such insights can tell us when the obtained saliency maps of trained networks point to wrong cues for classification (see Section \ref{sec:ablation_gender} for examples.) 
We now describe our methodology.
\section{Canonical Saliency Maps: Methodology}
\label{sec_method2}

The key aim of our methodology is to create a visualization which highlights the discriminative parts of a face for a given task. 
Our method is based on the assumption that the discriminative importance of a part of an input image is proportional to the drop in confidence of the classifier when the part is occluded \cite{deconvnet}, however on a canonical face representation. Like other occlusion-based saliency map methods, given an image $I \in \mathbb{R}^{W_I \times H_I}$ and the coordinates $(i,j)$, the importance of a patch $(|i-x|<\frac{sz}{2}$ $\forall x<W_I$, $|j-y|<\frac{sz}{2}$ $\forall y<H_I )$ is given as follows:
\begin{equation}
	S_{i,j} = \phi(I,c) - \phi(I\odot B_{i,j},c)
	\label{eq:drop_in_confidence}
\end{equation}
\noindent where $\phi(I,c)$ is the confidence of class $c$ for image $I$ and $B_{i,j} \in \{0,1\}^{W_I \times H^I}$ is a mask such that: 
\begin{align}
	B_{i,j}[x][y] & = 0 \text{ if } |i-x|<\frac{sz}{2} \text{ and } |j-y|<\frac{sz}{2} \\
	& = 1 \text{ otherwise } 
\end{align}  
\noindent and $sz$ is the size of the patch, which is a hyperparameter.

\subsection{ Alignment to a `Canonical' Face}
In order to capture the finer details of the parts of an image a trained DNN model looks at, we compute our saliency map on a standard neutral frontal face image $F \in \mathbb{R}^{W_F \times H_F}$ called the \textit{canonical face}, which helps compare saliency maps on a standardized platform.

We find an one-to-one mapping between the input face image and the canonical face image by fitting a 3D modular morphable model (3DMMM) \cite{bfm} using the procedure used by PR-Net \cite{prnet}. In particular, we use a convolutional neural network to regress a UV positional map from the input image, which gives the depth for a set of fixed points on the UV map of the face. For details of this procedure, please see \cite{prnet}. Let $M\in \mathbb{R}^{N \times 3}$ be a set of $N$ 3D points representing the 3DMMM. We fit it on the input image $I$ and the canonical image $F$ to obtain the set of 2D points $M_I \in \mathbb{R}^{N \times 2}$ and $M_F \in \mathbb{R}^{N \times 2}$ as the projection of $M$ on $I$ and $F$ respectively. Thus, we have a 1:1 dense mapping of points from $I$ to $F$ such that $I[M_I[n,1]][M_I[n,2]]$ refers to the same facial feature as $F[M_F[n,1]][M_F[n,2]]$ $\forall n \in \{1.2,\cdots,N\}$.
\FIGcisprocedure

\subsection{Mapping Discriminative Areas}
The Canonical Image Saliency (CIS) map is generated by accumulating the drop in confidence at each point of the dense alignment matrix $M_I$ and recording it on the corresponding location of $F$ on an intermediate matrix $P^* \in \mathbb{R}^{W_F \times H_F}$ as follows:
\begin{equation}
	\begin{aligned}
	P^*_{M_F[n,1],M_F[n,2]} & = 	P^*_{M_F[n,1],M_F[n,2]} \\
	 & + S_{M_I[n,1],M_I[n,2]}\\
	         \forall n<N
	 \end{aligned}
\end{equation}
where $P^*_{M_F[n,1],M_F[n,2]}$ is the patch around the point $(M_F[n,1],M_F[n,2])$ on the heatmap $P$, and $S_{M_I[n,1],M_I[n,2]}$ is the drop in confidence in the patch around point $(M_I[n,1],M_I[n,2])$ calculated according to Equation \ref{eq:drop_in_confidence}.

\subsection{Density Normalization}
Note that an equi-spaced grid on a 3-dimensional face may not correspond to equi-spaced grid on a 2D projection of the face. For example, on a frontal face image, the points on the sides of the face may be more spatially concentrated due to the curvature of the face. The heatmap values in these regions will hence be higher due to the concentration. We hence introduce a normalization step that keeps track of the number of times a pixel on an image is occluded, when performing the occlusion heatmap on the mesh. Let $N \in \mathbb{R}^{W_F \times H_F}$ be a matrix which stores the count of times each pixel of $P^*$ was updated. The final CIS map is calculated as follows:
\begin{equation}
	P = P^* \oslash N
\end{equation}
\noindent where $\oslash$ represents element-wise division. Figure \ref{fig:densitynorm} shows the effect of density normalization on the CIS map. 

\FIGdensitynorm

\subsection{From Image Saliency to Model Saliency}
We now discuss how the CIS maps are used to understand facial features that are important across all images for a given model trained for a specific task (for e.g, the part of the face that may be important for gender recognition vs another part that may be important for age recognition). We call these Canonical Model Saliency (CMS) Maps, which are model-level saliency visualizations to highlight facial areas that influence the model across all test images.

Given a test set $D$ consisting of images $\{I_1, I_2, I_3, ...\}$ with variations in factors such as pose, lighting, or expressions, we consider the average CIS map across these test images as the CMS map, i.e. 
\begin{equation}
	V= \frac{1}{N} \sum_i P_i \qquad  \forall I \in D 
	\label{eqn:cms}
\end{equation}
\noindent where $P_i$ is the CIS map of $I_i \in D$. It is possible to combine the CIS maps in other ways, but we found that simple averaging worked well in practice for model-level analysis. Learning CMS maps in other ways could be an interesting direction of future work.
Furthermore, in practice, we observe that it requires only a few images to generate a stable CMS map for a complete trained model. This suggests that face networks consistently rely on a few facial features and the canonical visualizations are stable across images. This is shown in Figure \ref{fig:cms_ablation} where we see that the trends become obvious from the first random 100 images. After 1000 images, the CMS is practically unchanged with the addition of more images. 
\FIGcmsablation

Figure \ref{fig:qualitative} shows a comparison between occlusion heatmaps of \cite{deconvnet} and our CIS maps. Our methodology is summarized as follows:

\begin{algorithm}
	\caption{Canonical Image Saliency Map}
    \hspace*{\algorithmicindent} \textbf{Input:} 
    \begin{itemize}[nosep,leftmargin=0.5in,topsep=0pt]
    	\item input image $I$ of size $W_I\times H_I$ 
    	\item input mesh $M_I$ of size $N\times 3$ 
    	\item frontal image $F$ of size $W_F\times H_F$
    	\item frontal mesh $M_F$ of size $N\times 3$ 
    	\item model $\phi$: deep model to find saliency where $\phi(I,c)$ gives the confidence of I for class c 
    	\item target class $C$ of the input image I 
    	\item $sz$: size of occlusion square
    \end{itemize}
    \hspace*{\algorithmicindent} \textbf{Output:} 
    heatmap $P$ of size $W_F\times H_F$
	\begin{algorithmic}[1]
		\Procedure{CIS}{$I, M_I, F, M_F, \phi, C, sz$}
		\State $P \gets \{0\}^{W_F\times H_F}$
		\State	$N \gets \{0\}^{W_F\times H_F}$ 
		\State $fsz \gets fsz\times \frac{H_F}{H_I}$ 
		\For{$i \gets 0$ \textbf{to} $n$ }
		\State $I^* \gets I$ 
		\State $I^*[M_I[i,0]-\frac{sz}{2}:M_I[i,0]+\frac{sz}{2}][M_I[i,1]-\frac{sz}{2}: M_I[i,1]+\frac{sz}{2}] \gets 0 $ 
		
		\State $x_F, y_F \gets M_F[i,0], M_F[i,1]$\\
		
		\State $P[x_F-\frac{fsz}{2}:x_F+\frac{fsz}{2}][y_F-\frac{fsz}{2}:y_F+\frac{fsz}{2}] +=  \phi(I,C) - \phi(I^*,C) $  
		
		\State $N[x_F-\frac{fsz}{2}:x_F+\frac{fsz}{2}][y_F-\frac{fsz}{2}:y_F+\frac{fsz}{2}] += 1$ 
		\EndFor
		
		\State $P[N=0] \gets 0 $
		\State $N[N=0] \gets 1 $
		\State $P \gets P \oslash N $ 
		\State return $P$
		\EndProcedure
	\end{algorithmic}
\end{algorithm}

\subsection{Model Saliency for Non-classification Tasks}
CMS maps can be generated for any face model which has a measure of confidence associated with each input image. Our method can be adapted to non-classification models by defining an appropriate confidence function.  Here, we define the confidence function for two commonly-used face tasks: zero-shot recognition using nearest neighbour and face verification. 

\subsubsection{Zero-shot Face Recognition}
Here, the query image $q$ is assigned the label of the image from the training set whose features have the highest cosine similarity with the features of the query image \cite{sphereface}. We define the confidence of classification in this setting as follows:
\begin{equation}
    S_{q, c} = \frac {A.Q} {\lVert A \rVert \lVert Q \rVert}
    \label{eq:zeroshotconf}
\end{equation}
where $c$ is the ground truth label of $q$, $Q$ is the feature of $q$ and $A$ is the feature of the closest training set image with label $c$. This new confidence function can be used in place of the class confidence $\phi$ in Equation \ref{eq:drop_in_confidence}.

\subsubsection{Face Verification}
Here, a pair of face images is considered to have the same identity if the cosine similarity between their features is more than a threshold calculated on the training set \cite{sphereface}. We define the confidence in this setting as follows:
\begin{equation}
    S_{q_1, q_2, c} = c \times (\tau - \frac {Q_1. Q_2} {\lVert Q_1 \rVert \lVert Q_2 \rVert} ) 
    \label{eq:verificationconf}
\end{equation}
where $c \in \{-1,1\}$ is the verification ground truth label, $\tau$ is the verification threshold, and $Q_1$ and $Q_2$ are the features of the image pair $q_1$ and $q2$. Using this function, we generate an IMS map for each pair of images and calculate the CMS map using Equation \ref{eqn:cms}.

\section{Experiments and Results}
\label{sec:results}
We now present our comprehensive experimental results, that analyze the effectiveness of canonicalizing saliency maps for face processing tasks. First, we explore our saliency maps through visual examples in Section \ref{sec:qual}. Second, we objectively assess the ability of our visualization to highlight discriminative parts of the face in Section \ref{sec:quant}. Third, we present the results of a user survey which shows that the parts of the face highlighted by our algorithm are important for the human perception of facial attributes in Section \ref{sec:survey}. Finally, we present extensive ablation experiments and discussions on our method in Section \ref{sec:ablation}. Unless otherwise mentioned, our experiments are conducted using the VGG-Face pre-trained model \cite{VGGFace} based on the VGG-16 architecture \cite{VGG16}. We use a random subset of the CelebA dataset \cite{CelebA} consisting of 22,000 images (henceforth called CelebA-subset) for all our experiments. (Note that these images are only used in the model's test phase, the model by itself is trained on all the training images in the CelebA benchmark). See the Supplementary Section for more details. 

\subsection{Qualitative Results}
\label{sec:qual}
\FIGqualitative
We compare the saliency maps produced by various methods in Figure \ref{fig:saliency_related}. As in Section \ref{sec:related}, most visualizations are not practically useful, and highlight a vague central portion of the face. 
In Figure \ref{fig:qualitative}, we display the visualization methods introduced in this work. From simple occlusion maps in column (a), we obtain Canonical Image Saliency (CIS) maps by projecting the occlusion maps onto a neutral frontal face, as shown in column (b). This `canonicalizing' allows us to collate the CIS maps to create Canonical Model Saliency (CMS) maps as shown in column (c). In column (d), we show that when the CMS maps are reprojected onto the input images, the saliency maps become meaningful for analysis.


\subsubsection{Evaluation of Canonical Model Saliency Maps on Various Face Tasks}
For this experiment, we used our algorithm on five models trained for the tasks of classification, expression, head pose, age and gender. We used the VGG-Face \cite{VGGFace} pre-trained model, and finetuned it for each of the aforementioned tasks on the CelebA \cite{CelebA} dataset. The ground truth labels for gender are provided with the CelebA dataset. The head pose ground truth was obtained by using PRNet \cite{prnet}, and the age ground truth was obtained using the DEX method \cite{agesota}. For expression, the ground truth for CelebA was obtained from a model trained on the FER 2013 data set \cite{Emotion_CelebA}. Since both head pose and age are real-valued, we grouped the values into discrete bins to convert them into classification tasks. For pose, the yaw and pitch values were binned into 9 bins ranging from top-left to bottom-right (see Figure \ref{fig:pose_nose}). Similarly, the real-valued ages obtained from the DEX model were grouped into 10 bins, each having 10 years. More details of the networks used are given in the Supplementary Section \ref{sec:supp_modeldetails}. The generated CMS maps are shown in Figure \ref{fig:cms}. We notice how models of the same architecture trained on different tasks focus on different face areas. For recognition, the eye-nose triangle is important and there is less focus on the mouth or the chin. Gender models surprisingly find the corners of the eyes to be the most discriminative facial features. We discuss the implications of this in Section \ref{sec:ablation_gender}. The nose is a crucial feature for the head pose model and the area between the eyebrows for the expression model. The age model looks at many different facial features. We see that CMS maps are a valuable asset to understand the nature of face tasks and the characteristics of various deep models when addressing these tasks. We discuss some of these results in more detail in Section \ref{sec:ablation}.


\subsubsection{Canonical Model Saliency Maps on Non-classification Tasks}
\FIGnoclassification
In this experiment, we show that CMS maps can be generated for non-classification face tasks. We generated CMS maps for zero-shot learning of face identities using nearest neighbour and face verification of VGG-Face fc1 features on the LFW \cite{lfw} dataset. For the zero-shot learning task, we occluded parts of the query image while using Equation \ref{eq:zeroshotconf} as the confidence function. For the verification task, we occluded the same region of both images in a verification pair and used Equation \ref{eq:verificationconf} as the confidence function. The results are shown in Figure \ref{fig:noclassification}. In both cases, we see the highlighted facial areas are similar to the classification task of recognition in Figure \ref{fig:cms}.

\subsubsection{Sanity Check Using Randomization}
\label{sec:sanity}
\FIGsanitycheck
\cite{sanitycheck} proposed a sanity check for saliency maps, where the layers of a trained model are progressively randomized starting from the output layer, and the changes in generated saliency maps are observed. A method is said to pass the sanity check if progressive randomization increases the randomization of the corresponding visualization. We perform a sanity check on our visualization using the same procedure, and reports the results of this experiment in Figure \ref{fig:sanity}. We observe that as more layers get randomized, the visualization gets more randomized. Thus, our method passes the sanity check. 

\subsection{Quantitative Results}
\label{sec:quant}
We conduct an objective evaluation of the faithfulness of our method on two datasets: CelebA and LFW \cite{lfw} and compare it with three popular saliency visualizations: GradCAM \cite{gradcam}, GradCAM++ \cite{gradcam++} and ScoreCAM \cite{scorecam}. Similar to \cite{gradcam++, scorecam}, we measure the confidence drop of explanation images produced by pixel-wise multiplication of the saliency heatmap with the base image. In particular, we utilize a `negative explanation image' by darkening the relevant areas of the base image. Unlike the task of object recognition, face images have a single object at the center of the image, and models trained on face images focus on different parts of the face image. In this process, saliency maps at times fail to detect the face completely (see Figure \ref{fig:metric}). Using negative explanation maps addresses such concerns. The negative explanation image $E$ is given by:
\begin{equation}
    E = (1-H) \otimes I
\end{equation}
where $H$ is the heatmap, $I$ is the base image and $\otimes$ represents pixel-wise multiplication. The heatmaps are first normalized to a range of [0,1] and the heatmaps for all the methods are standardized to have the same sum of pixels for each image:
\begin{equation}
    H' = \frac{h-\min(h)}{\max(h)-\min(h)}; 
    H = \frac{s}{\Sigma H'} H'
\end{equation}
where $h$ is the original heatmap, $s$ is a scalar which is the same for all heatmaps of the same image, and $H$ is the final heatmap which is used to create negative explanation maps. Normalizing the heatmaps in this way ensures that no visualization method gets an advantage of highlighting a large area of the input image, as only the discriminative parts should be highlighted. 

\FIGmetricjust

We adopt the three metrics used in \cite{gradcam++} with negative explanation images:

\noindent \emph{Average Drop \%:}
The confidence of an image when passed through a model is expected to decrease when the most discriminative parts are covered. We measure the drop of confidence when compared to the unmodified image as: 
\begin{equation}
 \frac{1}{N} \sum_{n=1}^N  {\max(0,\frac{ M(I_n) - M(E_n)}{M(I_N)}} \times 100
\end{equation}
where $M(E_n)$ and $M(I_n)$ are the confidence values of the $n^{th}$ explanation image and original image respectively. A high value of Average Drop \% indicates that the heatmap accurately highlights the most discriminative parts of the image.\\

\noindent \emph{\% Increase in confidence:}
In some images, covering the highlighted parts may result in an undesired increase in confidence with respect to the original image. We measure the number of such images using this measure as follows:
\begin{equation}
\frac 1 N \sum_{i=1}^N \mathbb{I}_{M(E_n) > M(I_n)}\times 100
\end{equation}
where $\mathbb{I}$ is the indicator function which returns 1 if $M(E_n)>M(I_n)$ and 0 otherwise. A low score in this metric is better.\\ 

\noindent \emph{Win \%:}
Here, we compare all the four methods and measure which method produces the greatest drop in confidence for a given test image. For example, Win \% of CMS is calculated as follows:
\begin{equation}
\begin{split}
    \frac{1}{N} \sum_{i=1}^N \mathbb{I}_{M(E^{CMS}_n) < (M(E^{GradCAM}_n),} \qquad \qquad \qquad \\
    \qquad _{M(E^{GradCAM++}_n), M(E^{ScoreCAM}_n))} \times 100
\end{split}
\end{equation}
where the indicator returns 1 if the explanation map produced by CMS has the lowest confidence. The sum of Win \% across all the visualization methods for a single task should add up to 100. 

We conduct three experiments for quantitative evaluation. First, we calculate the above metrics on VGG-16 for the tasks of recognition, gender, age, head pose and expression on the CelebA dataset.
For fair comparison, we use our maps projected back onto the input image (Col (d) of Figure \ref{fig:qualitative}).
Figure \ref{fig:vggfacequant} shows our results and a comparison with other visualization methods. Our method outperforms all other methods in all metrics. The Win \% shows that for most images, removing the explanation map given by our method causes the highest drop in confidence (higher the better). 

\FIGVGGFaceQuant

Secondly, we repeat the experiment on the LFW \cite{lfw} dataset using the VGG-Face network, using the same experimental settings as above. We show the results in Figure \ref{fig:lfwquant}. Here too, our method outperforms all other methods by a large margin in all quantitative metrics, showing that our method generalizes across datasets. 

\FIGlfwquant

We also compare our saliency methods on various off-the-shelf gender models. We use pretrained models from \cite{fairface, agesota, cpg} and evaluate our metrics on CelebA-subset. More details about these models are given in the Supplementary Section \ref{sec:supp_modeldetails}. Our results are shown in Figures \ref{fig:netquant}. Once again, we see that our method outperforms all other methods on all metrics.  We show the CMS maps obtained using the various networks in Figure \ref{fig:netcms}.

\FIGnetquant
\FIGnetcms

\subsection{User Survey on Perception of Facial Attributes}

\label{sec:survey}
We conducted a user survey to evaluate the human interpretability of our saliency maps as compared to other visualization methods. In particular, we explored whether the discriminative facial areas found by Canonical Model Saliency Maps are vital for human perception of facial attributes. We focused on the tasks of gender and expression for this study. The survey used a total of 96 images, each of which were evaluated by 154 participants not involved in this work. Twelve base images for each task were used, for which we generated four negative explanation maps corresponding to the four saliency visualization methods GradCAM, GradCAM++, ScoreCAM and reprojected CMS maps using the Gender and Expression models mentioned in Section \ref{sec:quant}. We also applied a vignette to each of the explanation images to hide the context information (see Figure \ref{fig:survey} for sample images). Each participant was given a binary choice for each image (male-female or happy-sad, depending on the task). Since a better visualization algorithm hides crucial information and makes it more difficult to interpret an image, we use the percentage of wrong answers as a measure of the goodness of the visualization method. We show some sample survey images in Figure \ref{fig:survey_samples}. See the Supplementary section for more examples.
\FIGsurveysamples
The results of our survey are given in Figure \ref{fig:survey}. We see that the percentage of wrong answers marked by the respondents is higher for our method than other methods, indicating that our method performed better at hiding the most crucial and discriminative facial areas.
\FIGSurvey

\section{Analysis and Discussion}
\label{sec:ablation}
In this section, we present analysis of the proposed method including ablation studies and discussions. 

\subsection{Why Model-level Saliency Maps?}
\FIGvaluealignment
Canonical Model Saliency (CMS) maps allow us to observe patterns and trends in the functioning of deep face models by adding the simple yet powerful step of alignment of occlusion-based saliency maps to a canonical face model. For example, using CMS maps, we observed that the corners of the eyes are important for gender classification (Section \ref{sec:ablation_gender}). This is not directly apparent by observing individual, unaligned occlusion maps, as seen in Figure \ref{fig:value_alignment}. The advantage of this alignment process is in allowing comparison and aggregation of saliency maps. 
A single occlusion map may contain variations caused by differences in the image setting such as pose, occlusion and lighting, thus not allowing us to understand the whole picture. The process of aggregation averages out the effects of variations in individual images, showing us the parts of the face that are truly important. 

\subsection{Effect of Make-up on Gender Classification}
\label{sec:ablation_gender}
\FIGgenderablation
\FIGsizeablation
The CMS maps for the gender model provided interesting insights using our method (Figure \ref{fig:cms}E). We expected the heatmap to highlight the areas around the mouth, jaw and cheeks, as they contain facial hair cues and different bone structure for different genders. However, the map showed that the model fixated mostly on eye corners. We hypothesize that this is because the model was finetuned on the CelebA dataset \cite{CelebA}, which consists of images of celebrities who use make-up extensively. The model picked up on the cue of eye make-up to classify gender. We presume that such a model will not work well for a different demographic distribution. This may be the reason why many commercial face models fail in detecting gender for females and different races \cite{gendershades}. This indicates the importance of detecting dataset biases as they can have a significant impact on the performance of deep models.
We test our hypothesis with the following qualitative experiment. We collect a few images of people with and without eye make-up from the Internet. These images were passed through the gender model and the confidence for `male' and `female' classification was observed. Our results are presented in Figure \ref{fig:gender_ablation_big}. We observed that in all cases, there was a drop of confidence in `male' classification when the men wore make-up and a smaller drop in confidence of `female' classification for women without make-up. In some cases, the drop in confidence was large enough to flip the original classification result. This was especially true for males of Asian origin, especially those from the far East. We conclude that eye make-up has a significant effect on the performance of such a gender model, which is skewed towards people of a certain ethnicity.

\subsection{Head Pose Model Relies on the Nose}
\FIGposenose
The shape of the nose changes according to the pose of the face (Figure \ref{fig:pose_nose}A). Generally, the nose is positioned at the centre of the face, and its placement on the face changes consistently with the 3D orientation of the face. The head pose can be detected quite accurately from the shape of the nose and the quadrant of the face in which the nose tip resides (along with the jawline), especially when there are only nine classes, as shown in Figure \ref{fig:pose_nose}. The nose thus provides the strongest cue for the head pose. This is reflected in the CMS map shown in Figure \ref{fig:cms}D.

\subsection{Age Model Uses the Whole Face}
The CMS map for age (Figure \ref{fig:cms}F) shows that the cues for age are present in multiple areas of the face. Some of the distinctive features for age may be the tightness of skin around the eyes and jaws, wrinkles and receding hairline. Pre-deep learning methods used the geometry or texture of the face for age prediction \cite{agesurvey}, thus corroborating our finding on why age-related cues are found all over the face. 

\subsection{How Occlusion Size Affects Saliency Maps}
We present a qualitative ablation study to explore the effect of the size of the occluding patch on the generated CIS map. The number of vertices provided by the dense face alignment algorithm is very high and the time required to compute the heatmap at each vertex is large. Hence we use a tunable `stride' parameter to omit vertices at regular intervals. As the size of the occluding patch decreases, a smaller stride is chosen so that gaps don't appear in the visualization. The stride can be larger for bigger occluding patches without affecting the visualization quality. In Figure \ref{fig:size_ablation}, we show the result of changing the patch size on the CIS maps generated from the same input image. We observe that as the patch size increases, the map becomes fuzzier but general patterns do not change. Our method provides useful information regardless of the size of the occluding patch, although smaller patches give better resolution. We used a patch of size $15\times 15$ for generating other saliency maps in this work, as it provides a good balance between heatmap resolution and computation time.

\subsection{Why Align to Canonical Face?}
\label{sec:align_ablation}
\FIGalignmentablation
\FIGalignmentqual
Here we examine the need for a canonical face instead of using keypoint-based alignment or the image pixel positions. The main advantage of canonical face alignment is that it ensures that the model saliency maps remain accurate while aggregating the individual image saliency maps. If we do not align the heatmaps precisely, the changes in position add up to produce an inaccurate model saliency map. 

We conduct an ablation study to demonstrate this effect. We use three types of alignment and generate model saliency maps on the LFW dataset: 1) no alignment; 2) keypoint-based alignment; and 3) canonical face alignment. For the first case, we create image saliency maps by sliding an occlusion window over the entire input image. We repeat the procedure for the second case, but we used LFW images aligned with keypoint alignment \cite{lfwfunneled} as the input instead of the raw LFW images. The third case used the same setting as previous CMS experiments. 
We create the model saliency maps for each case by averaging individual image saliency maps. We generate explanation maps and calculate quantitative metrics. The results are shown in Figure \ref{fig:alignment_ablation}. Canonical alignment performs better than keypoint-based alignment or no alignment in all cases. We show all three model saliency maps in Figure \ref{fig:alignment_qual}. 

Using canonical faces also results in lower computation cost, as we know exactly which parts of the image we need to occlude, as opposed to sliding the occlusion patch over the whole image. 

\subsection{Robustness in Deep Models}
Robustness refers to the property of a model wherein small deviations in input images, due to noise or natural variations, do not affect the correctness of the model. If a model relies on a small set of cues, it is more likely to go wrong due to input image diversity. Instead, if the model looks at many cues, small variations are less likely to confuse the model. The CMS maps indicate the areas from which deep models pick up cues. The maps thus also allow us to obtain an estimate of the model's robustness. A model that concentrates on a few facial areas is likely to be less robust than one that focuses on many facial areas. Less robust models are more prone to mistakes when presented with extreme cases of occlusion, lighting and other deviations. We see an example with our trained gender model (Section \ref{sec:ablation_gender}), where the model is not robust to changes in the face due to make-up.

\section{Conclusion}

In this work, we showed that standardization of saliency maps via Canonical Saliency Maps provides usable and interpretable results in the face domain when compared to current saliency methods which give trivial outputs for face images. Canonical Saliency Maps highlight the facial areas of importance by projecting occlusion-based heatmaps onto a neutral face. Computing model-level canonical saliency maps enable us to perceive which facial features are important for different face tasks, thereby revealing the strengths and weaknesses of face models. These observations can be compared to human perception, which can show us if the model is behaving in unexpected ways. The maps aid in detecting problems and biases inherent in the model. In particular, by utilizing Canonical Model Saliency maps, we identified a bias in a gender model, wherein the model was wrongly using make-up as a cue to classify gender. We confirmed the presence of the bias with additional studies. Such models can cause problems when used in demographics unlike the training dataset, where the patterns of applying make-up are different. 

Nowadays, deep face models are deployed in critical applications like security and law enforcement -- the proposed Canonical Saliency Maps allow such systems to be critically analyzed before deployment, and thus increase trust. They can also be used to predict failures during development and help improve the models. We hope that the tools presented in this work, while simple, can be very effective in practical use for deeper understanding of face models, their biases and failures. 
In future work, we aim to study methods of mitigating the problems and biases detected by our visualization methods.
\ifCLASSOPTIONcompsoc
\section*{Acknowledgments}
\else
\section*{Acknowledgment}
\fi

VNB would like to thank MHRD, Govt of India, and Honeywell for funding support through the UAY program (UAY-IITH005).

\newpage

\bibliographystyle{IEEEtran}
\bibliography{IEEEabrv,egbib}
%

\begin{IEEEbiography}[{\includegraphics[width=1in,height=1.5in,clip,keepaspectratio]{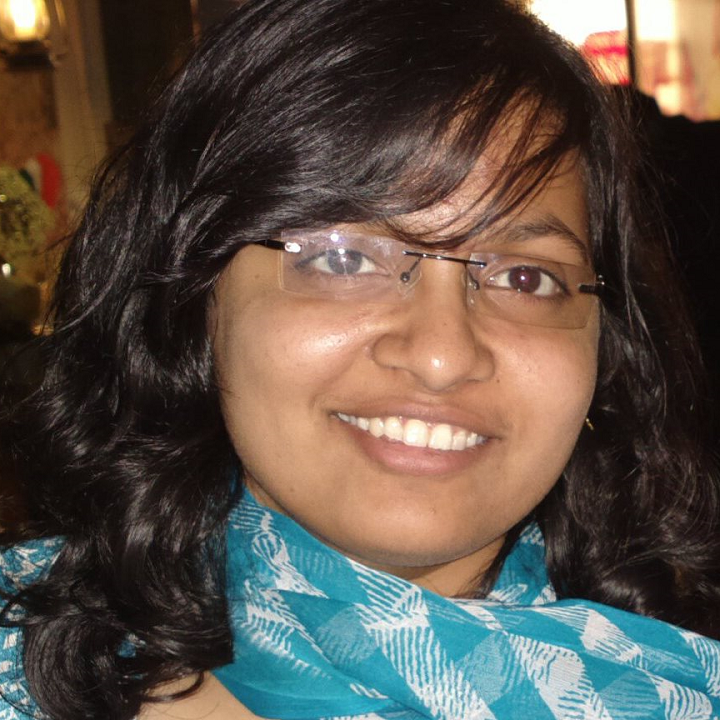}}]{Thrupthi Ann John} recieved the B.Tech in Computer Science from the National Institute of Technology, Warangal. She is currently a PhD student in the Computer Vision and Information Technology lab in IIIT Hyderabad. She is advised by Prof C V Jawahar and Prof Vineeth N Balasubramanian. Her interests include deep algorithms for face tasks, visualization of deep algorithms and solving computer vision problems using deep learning. 
\end{IEEEbiography}

\begin{IEEEbiography}
[
{\includegraphics[width=1in,height=1.5in,clip,keepaspectratio]{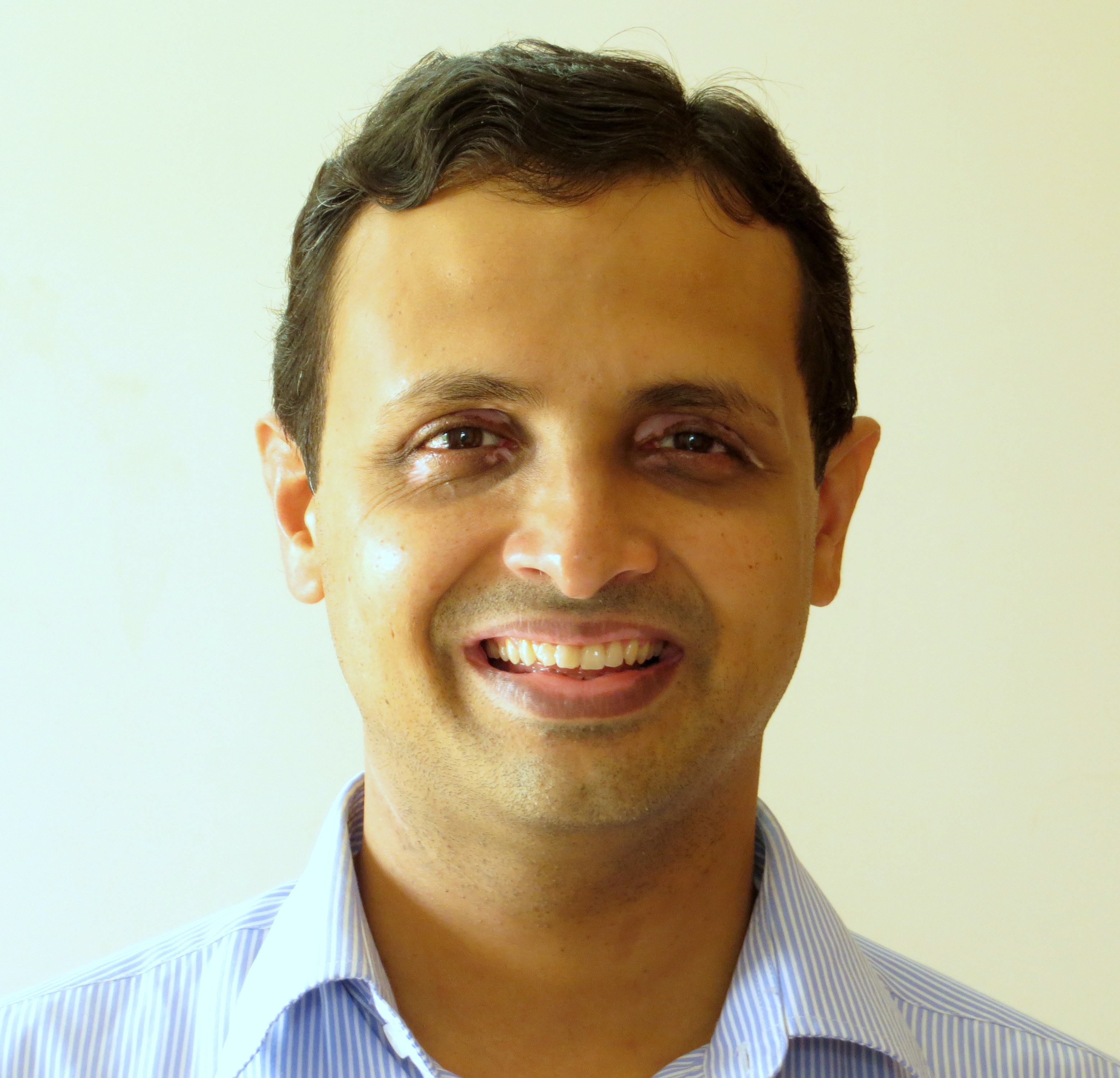}}
]%
{Vineeth N Balasubramanian}
is an Associate Professor in the Department of Computer Science and Engineering at the Indian Institute of Technology, Hyderabad (IIT-H). His research interests include deep learning, machine learning, and computer vision, with a focus on explainable deep learning and learning with limited supervision. His research has been published at premier peer-reviewed venues including ICML, NeurIPS, CVPR, ICCV, KDD, ICDM, IEEE TPAMI and IEEE TNNLS. For more details, please see \url{https://iith.ac.in/~vineethnb/}.
\end{IEEEbiography}\

\begin{IEEEbiography}[{\includegraphics[width=1in,height=1.5in,clip,keepaspectratio]{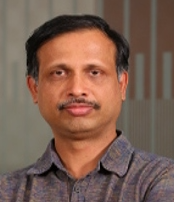}}]{C V Jawahar}
is  the Amazon Chair Professor at IIIT Hyderabad, India, where he leads a group focusing on AI, computer vision, and machine learning. His primarily research interest is in a set of problems that overlap with vision, language and learning. Prof Jawahar is a Fellow of INAE and IAPR.
\end{IEEEbiography}




\cleardoublepage

\section*{Supplementary Section}
\beginsupplement

In this section, we provide additional details that could not be added in the paper due to space constraints. The contents are as follows:
\begin{enumerate}
    \item Details of the models used for experiments in the paper
	\item Details of user survey

\end{enumerate}

\section{Details of Models}
\label{sec:supp_modeldetails}

Our main experiments are conducted on five different models trained for five tasks. The details of these models are given in Table \ref{tab:model_details}. Models for Expression, Pose, Gender and Age were obtained by finetuning the VGG-Face model on the Celeb-A \cite{CelebA} dataset. The ground truth for gender was provided in the dataset. For age, emotion and pose, we generated the ground truth using known methods. The ground truth for age was obtained using the method DEX: Deep EXpectation of apparent age from a single image \cite{agesota}. This method uses a VGG16 architecture and was trained on the IMDB-WIKI data set which consists of 0.5 million images of celebrities crawled from IMDB and Wikipedia. The ages obtained using this method were binned into 10 bins, with each bin having 10 ages. Head pose was obtained by registering the face to a 3D face model using linear pose fitting \cite{eos_headpose}. The model used is a low-resolution shape-only version of the Surrey Morphable Face Model. The yaw and pitch values were binned into 9 bins ranging from top-left to bottom-right. The binned pose values are shown in Figure \ref{fig:pose_nose}. For emotion, the ground truth was obtained using a VGG-16 model trained on FER 2013 data set \cite{fer2013} with 7 classes. The accuracy for the recognition models are reported on the LFW dataset. The accuracy for other models are reported on a test partition of the Celeb-A dataset.
We used three additional models for our experiments on gender detailed in Sections \ref{sec:quant}. The details of these models are given in Table \ref{tab:gendermodels}.

\begin{figure}[!ht]
	\centering
	\includegraphics[width = \linewidth]{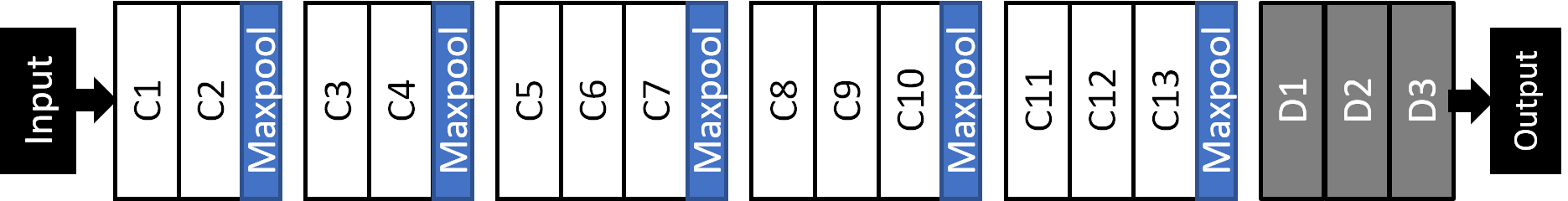}
	\caption{Most experiments in this work are conducted on the VGG-Face network \cite{VGGFace} shown above,  which follows the architecture of VGG-16 \cite{VGG16}. Layers C1 to C13 stand for convolutional layers. Layers D1, D2 and D3 represent fully connected dense layers. There is a ReLU non-linearity after each convolutional and fully connected layer}
	\label{fig:vggface}
\end{figure}

\begin{table}[!ht]
    \centering
    \begin{tabular}{|l|l|l|l|}
    \hline
        \textbf{Task} & \textbf{Architecture} &\textbf{Training} & \textbf{Accuracy} \% \\
        \hline \hline
        Recognition & VGG-16 \cite{VGG16} & VGG-Face \cite{VGGFace}& 98.95 on LFW \cite{lfw}\\ \hline
        Recognition & LightCNN-9 & \begin{tabular}{@{}c@{}}Casia-WebFace \\ MS-Celeb-1M\end{tabular} & 98.8 on LFW \\ \hline
        Expression & VGG-16 & \begin{tabular}{@{}c@{}}Celeb-A using  \\ FER13 \cite{fer2013}\end{tabular} & 69.01 \\ \hline
        Pose & VGG-16 & \begin{tabular}{@{}c@{}}Celeb-A using  \\3DMM \cite{eos_headpose}\end{tabular}  & 96.62\\ \hline
        Gender & VGG-16 & Celeb-A \cite{CelebA}& 98.37 \\ \hline
        Age & VGG-16 & \begin{tabular}{@{}c@{}}Celeb-A using  \\IMDB-Wiki \cite{agesota}\end{tabular} &  61.72\\
        \hline
    \end{tabular}
    \caption{Details of deep face models used in this work}
    \label{tab:model_details}
\end{table}

\begin{table}[!ht]
	\centering
	\begin{tabular}{|l|l|l|}
		\hline
		Model name & Implementation & Base Architecture \\
		\hline \hline
		VGG Gender & Trained by authors on CelebA & VGG-16 \\
		Fairface \cite{fairface} & https://github.com/dchen236/FairFace & resnet34\\
		DEX \cite{agesota} & https://github.com/siriusdemon/pytorch-DEX & VGG-16 \\
		CPG \cite{cpg} & https://github.com/ivclab/CPG & spherenet \\
		\hline
	\end{tabular}
	\caption{Details of the deep gender models used for Figures \ref{fig:netquant} and \ref{fig:netcms}}
	\label{tab:gendermodels}
\end{table}
\section{Details of User Survey}
\FIGsurveybase

In this section, we give additional details on the user study conducted to understand human perception of facial expressions in Section \ref{sec:survey}. The study was conducted to determine the parts of the face that are important for human perception of expressions and compare it to machine perception. All the base images used in our survey are given in Figure \ref{fig:survey_base}. There were four negative explanation images created from each base image by using GradCAM, GradCAM++, ScoreCAM and CMS, as shown in Figure \ref{fig:survey_samples}. The survey was conducted using Google Forms. The 64 images were arranged into four pages such that each page contained one variation of each base image. The variations were mixed equally across the four pages.


\ifCLASSOPTIONcaptionsoff
  \newpage
\fi

\end{document}